\pdfoutput=1

\documentclass[11pt]{article}

\usepackage[final]{acl}

\usepackage{times}
\usepackage{latexsym}

\usepackage[T1]{fontenc}

\usepackage[utf8]{inputenc}

\usepackage{microtype}

\usepackage{inconsolata}

\usepackage{graphicx}

\graphicspath{{assets}}

\newif\iflowres
\lowresfalse

\newcommand{\myincludegraphics}[2][]{%
  \iflowres
    \includegraphics[#1]{#2_lowres}%
  \else
    \includegraphics[#1]{#2}%
  \fi
}

\usepackage{booktabs}
\usepackage{amsmath}
\usepackage{amssymb}
\usepackage{multirow}
\usepackage{multicol}
\usepackage{xcolor}
\usepackage{colortbl}
\usepackage{eucal} %
\usepackage{enumitem} %
\usepackage{bm} %
\usepackage{mdframed}

\newcommand\sX{\ensuremath{\mathcal{X}}}
\newcommand\R{\ensuremath{\mathbb{R}}}

\newcommand{\argmin}{\mathop{\mathrm{arg\,min}}}

\newcommand{\pretraindatasize}{n_{\text{pt}}}
\newcommand{\finetunedatasize}{n}
\newcommand{\inferencedatasize}{n_{\text{inf}}}
\newcommand{\computebudget}{c}
\newcommand{\datascalingexponent}{d}

\newcommand{\greencheck}{{\color{green}\checkmark}}
\newcommand{\redcross}{{\color{red}$\times$}}

\newcommand{\importantstatement}[1]{%
    \begin{mdframed}[leftline=true,linewidth=.5pt,linecolor=black!100]
        #1
    \end{mdframed}
}

\title{
    Inference Compute-Optimal Video Vision Language Models
}

\author{
 \textbf{Peiqi Wang\textsuperscript{1},}
 \textbf{ShengYun Peng\textsuperscript{2},}
 \textbf{Xuewen Zhang\textsuperscript{3},}
 \textbf{Hanchao Yu\textsuperscript{3},}
 \textbf{Yibo Yang\textsuperscript{3},}
\\
 \textbf{Lifu Huang\textsuperscript{4},}
 \textbf{Fujun Liu\textsuperscript{3},}
 \textbf{Qifan Wang\textsuperscript{3}}
\\
\\
 \textsuperscript{1}MIT,
 \textsuperscript{2}Georgia Tech,
 \textsuperscript{3}Meta,
 \textsuperscript{4}UC Davis
\\
 \small{
   Email: \href{wpq@mit.edu}{wpq@mit.edu}\quad Code: \href{https://github.com/tt6746690/vvlm_inference_scaling}{github/vvlm\_inference\_scaling}
}   
}

\begin{document}
\maketitle

\begin{abstract}
This work investigates the optimal allocation of inference compute across three key scaling factors in video vision language models: language model size, frame count, and the number of visual tokens per frame. 
While prior works typically focuses on optimizing model efficiency or improving performance without considering resource constraints, we instead identify optimal model configuration under fixed inference compute budgets.
We conduct large-scale training sweeps and careful parametric modeling of task performance to identify the inference compute-optimal frontier. Our experiments reveal how task performance depends on scaling factors and finetuning data size, as well as how changes in data size shift the compute-optimal frontier. These findings translate to practical tips for selecting these scaling factors.
\end{abstract}

\section{Introduction}
\label{sec:introduction}

Consider a real-world scenario where we are finetuning a video vision language model (video VLM) for deployment to process millions of videos each day. In this case, the pretraining compute cost has already been absorbed by other parties (e.g., via open source), and the cost of finetuning is negligible compared to the massive scale of video processing demand required, as Appendix~\ref{sec:appendix_faq_finetuning_compute_cost_is_negligible} explains in more detail. Consequently, inference compute cost dominates operational considerations.

Important design decisions $x$ made before finetuning can significantly impact both the model’s inference compute cost $c(x)$ and downstream task performance $f$. For video VLMs, the most important parameters include: (1) the language model (LM) size $x_N$, (2) the number of frames processed per video $x_T$, and (3) the number of tokens per frame $x_V$ used to represent videos. Once these parameters are selected, the model is finetuned on the available training data and then deployed under the same setup $x$.

Given the substantial computational challenges of deploying video VLMs at scale, our work addresses a key question:
\importantstatement{\textit{\textbf{Q}: Given fixed finetuning data of size $\finetunedatasize$ and per-example inference compute budget $\computebudget$, how to select scaling factors $x$ for the best model?}}
We focus on three specific scaling factors $x \triangleq (x_N,x_T,x_V)$, although our framework can extend to any scaling factors that affects inference compute cost and performance. This problem persists across different pretrained models, inference acceleration techniques, training data characteristics, and evaluation metrics. While specific inference scaling behaviors may vary depending on these factors \citep{zhangWhenScalingMeets2023}, we provide a reasonable approach to tackle this problem and derive transferable insights.

Concurrent studies exploring scaling trade-offs in vision-language models have notable gaps \citep{liInferenceOptimalVLMs2024,duExploringDesignSpace2024}. For instance, \citet{duExploringDesignSpace2024} examines the trade-off between frame count $x_T$ and tokens per frame $x_V$ but overlooks other important scaling factors like language model size $x_N$. Moreover, it does not account for the computational cost of processing video frames through vision encoders, leading to overestimated benefits of increasing the number of frames $x_T$. Our work addresses these limitations.

Existing research on scaling inference-time compute \citep{liInferenceOptimalVLMs2024,wuInferenceScalingLaws2024,snellScalingLLMTestTime2024,brownLargeLanguageMonkeys2024} don't investigate the possible interaction between scaling factors $x$ and finetuning data size $\finetunedatasize$. Scaling factors influence how effectively finetuning data is used: (1) larger models are more sample-efficient, achieving lower error with the same amount of data \citep{henighanScalingLawsAutoregressive2020}, and (2) increasing the number of frames or tokens per frame enriches the dataset by providing more detailed visual information. In contrast, we conduct experiment to explore these interactions and derive insights into how finetuning data size $\finetunedatasize$ shifts the compute-optimal frontier. 

In this paper, we address the inference compute allocation problem for video VLMs and propose a simple, effective recipe. Inspired by \citet{hoffmannTrainingComputeoptimalLarge2022,alabdulmohsinGettingViTShape2023}, we conduct efficient training sweeps to explore multiple scaling factors (e.g., $x_N$, $x_T$, $x_V$, $\finetunedatasize$) while minimizing the number of training runs. To address the lack of consensus on appropriate parametric models for downstream task performance \citep{owenHowPredictableLanguage2024,liInferenceOptimalVLMs2024,gadreLanguageModelsScale2024}, we perform model selection to identify suitable formulations and thoroughly validate the fitted model, accounting for challenges such as small sample sizes and the difficulty of predicting performance metrics (vs. next-token prediction loss). Finally, we determine the optimal scaling factors by solving a constrained discrete optimization problem numerically, providing a concrete answer to the posed question.

Our training sweeps reveals that: (1) scaling factors $x$ and finetuning data size $\finetunedatasize$ exhibit diminishing returns on performance, (2) jointly scaling $(x_N, x_T, x_V)$ is crucial for optimal performance, and (3) the utility of $x$ varies significantly across tasks, indicating that no universal inference compute allocation strategy exists—allocation must be task-specific. We demonstrate that a power-law additive functional form with an interaction term effectively models average task performance. By solving a constrained optimization problem using this parametric model, we identify an inference compute-optimal frontier that provides guidance on scaling $x$ optimally. Moreover, the finetuning data size $\finetunedatasize$ shapes this frontier; for instance, the trend suggests increasing $x_T$ and $x_V$ while decreasing $x_N$ as more data becomes available.

In summary, our key contributions are:
\begin{itemize}[noitemsep, left=0pt]
    \item We tackle the important problem of allocating inference compute across scaling factors in video VLMs. Our approach combines efficient training sweeps, parametric modeling of task performance, and constrained discrete optimization to identify optimal trade-offs.  
    \item We conduct large-scale training sweeps (e.g., $\sim$100k A100 hours) and derive qualitative observations of how video task performance varies with scaling factors and finetuning data size.
    \item We provide actionable insights into scaling inference compute for video VLMs, e.g., optimal allocation of factors such as LM size, frame count, and tokens per frame, as well as the impact of data size on the compute-optimal frontier.
\end{itemize}

\section{Relationship to Scaling Law Studies}
\label{sec:relationship_to_scaling_law_studies}

In contrast to previous studies on training compute-optimal model that balances model size $x_N$ and pretraining data size $\pretraindatasize$ \citep{kaplanScalingLawsNeural2020,hoffmannTrainingComputeoptimalLarge2022}, we aim to optimize scaling factors for video VLMs such that the model is inference compute-optimal. While our motivation aligns with \citet{sardanaChinchillaOptimalAccountingInference2024} in considering deployment costs, our focus is on finetuning rather than pretraining LMs. A notable distinction of our setup is that we finetune with all available data because finetuning datasets are limited in quantity, their compute cost is negligible, and more data always provide better performance without incurring inference compute cost. Table~\ref{tab:compare_train_vs_inference_compute_optimal_scaling_study} contrasts the optimization of training compute for LMs and inference compute for video VLMs.

\begin{table}[t]
  \centering
  \begin{tabular}{lll}
  \toprule
  \textbf{Aspect} & \textbf{Training} & \textbf{Inference} \\
  \midrule
  Data & Abundant & Limited  \\
  Cost ({\scriptsize FLOPs}) \textdagger  & $6 x_N \pretraindatasize$ & $2x_N x_T x_V$ \\
  Problem & $\min\limits_{x_N, \pretraindatasize} f(x_N,\pretraindatasize)$ & $\min\limits_{x} f(x,\finetunedatasize)$ \\
               & $c(x_N,\pretraindatasize) \leq \computebudget_{\text{pt}}$ & $c(x) \leq \computebudget$ \\
  \bottomrule
  \end{tabular}
  \caption{
    Comparison of optimizing training compute for LLMs versus inference compute for video VLMs. Notations: $x_N$ (LM parameters), $x_T$ (frames), $x_V$ (tokens per frame), $\pretraindatasize$ and $\finetunedatasize$ (training and finetuning data sizes), $f$ (error), $\computebudget_{\text{pt}}$ (training compute budget), and~$\computebudget$ (per-example inference compute budget). \textdagger denotes compute cost estimates for LMs only.
  }
  \label{tab:compare_train_vs_inference_compute_optimal_scaling_study}
\end{table}

Motivated by existing finetuning scaling studies, our work investigates how finetuning data size~$\finetunedatasize$ affects VLM performance. Unlike \citet{hernandezScalingLawsTransfer2021}, which focuses on pretraining-to-finetuning knowledge transfer, we explore how finetuning data size interacts with other scaling factors to influence inference efficiency and performance. While \citet{zhangWhenScalingMeets2023} examines similar interactions in the context of finetuning LMs for translation and summarization, we target video benchmarks and focus on allocating inference compute.

Our work can be viewed as scaling video VLMs' inference-time compute, e.g., by increasing the size of video representations. While previous research has explored scaling retrieval augmented generation \citep{yueInferenceScalingLongContext2024} or test-time search strategies (e.g., best-of-N or beam search) \citep{wuInferenceScalingLaws2024,snellScalingLLMTestTime2024,brownLargeLanguageMonkeys2024}, these approaches adjust inference compute for a fixed model. In contrast, we finetune models for any given scaling factors and use the same setup during inference.

\section{Inference Computational Optimality}

\paragraph{Notations}

Let $x \in \sX_1 \times \cdots \times \sX_K \triangleq \sX$ denote a set of $K$ factors that influence both inference compute and video VLM performance, where each $\sX_k \subset \mathbb{N}$ is a subset of the natural numbers. In this work, we specialize $x$ to $(x_N, x_T, x_V)$, where $x_N$ denotes the language model (LM) size, $x_T$ the number of video frames, and $x_V$ the number of visual tokens per frame, with $\sX_N$ restricted to pretrained LM sizes and $\sX_V$ to perfect squares. Model performance is quantified using a metric $f: \sX \times \mathbb{N} \to \mathbb{R}$, where $f(x, \finetunedatasize)$ represents the downstream task error (lower is better) from finetuning a model with scaling factors $x$ on a dataset of size $\finetunedatasize$. While~$f$ represents error, we refer to it interchangeably as performance throughout this paper.

\paragraph{Video VLM}

Our analysis of video VLM scaling is based on LLaVA-like architectures \citep{liuVisualInstructionTuning2023}, which consist of two main components: 
\begin{itemize}[noitemsep, left=0pt]
    \item \textit{vision model} with $x_M$ parameters (e.g., CLIP \citealp{radfordLearningTransferableVisual2021}) that processes $x_T$ video frames independently. For each frame, it generates a grid of $x_W$ visual features that are then projected and resampled into $x_V$ visual tokens
    \item \textit{language model} with $x_N$ parameters (e.g., Llama-3) that consumes the sequence of visual token representations of length $x_Tx_V$ to perform video understanding or reasoning tasks.
\end{itemize}

\paragraph{Inference Compute Cost}
\label{sec:method_inference_compute_cost}

We measure computational cost using floating-point operations (FLOPs), focusing on the inference cost of the vision and language model components. We assume each example consists of a single video, where the length of visual tokens dominate that of the input instructions or output generations. Thus, we disregard the compute cost of the latter. Using the standard approximation of $2x_N$ FLOPs per token for a transformer model with $x_N$ parameters \citep{kaplanScalingLawsNeural2020}, the per-example inference compute cost is
\begin{align}
    \label{eq:inference_compute_of_video_vlm_with_vit}
    c(x)
        = 2 x_T \left( x_M x_W + x_N x_V \right).
\end{align}
For a video VLM using SoViT-400m/14 \citep{alabdulmohsinGettingViTShape2023} as the vision model, this becomes $c(x) = 2 x_T ( 0.43e9 \cdot 768 + x_N x_V )$.

Prior work often overlooks the vision model's compute cost, focusing solely on the language model \citep{liInferenceOptimalVLMs2024,duExploringDesignSpace2024}. In Appendix~\ref{sec:appendix_faq_compute_cost_of_vit_matters}, we demonstrate that the vision model's relative compute cost becomes significant as $x_N$ and $x_V$ decrease.

\paragraph{Parametric Model of Performance}
\label{sec:method_parametric_model_of_performance}

We model task error $f(x, \finetunedatasize)$ using \texttt{add-interact} \citep{alabdulmohsinGettingViTShape2023}, a parametric function that defines an additive power-law relationship with interaction terms. This formulation is well-suited for scenarios where both inference compute $c(x)$ and data size $\finetunedatasize$ are constrained, as is often the case when deploying video VLMs. Moreover, it enables an exploration of the interactions between $x$ and $\finetunedatasize$.

For a specific scaling factor $x_k \in \sX_k$ (e.g., the number of frames $x_T$) and finetuning data size $\finetunedatasize$, we model the error as
\begin{align}
    \label{eq:add_interact_functional_form}
    f_k(x_k, \finetunedatasize)
        = \alpha_k x_k^{-a_k} + \left( \beta_k x_k^{b_k} + \xi_k \right) \finetunedatasize^{-\datascalingexponent} + \varepsilon_k,
\end{align}
where $\alpha_k, \beta_k, \xi_k, \varepsilon_k \geq 0$ are coefficient parameters, and $a_k, b_k, c \in \R$ are exponent parameters to be estimated from empirical training results. The terms in this formulation are interpreted as follows:
\begin{itemize}[noitemsep, left=0pt]
    \item The coefficients $\alpha_k, \beta_k, \xi_k$ represent error components that can be reduced by increasing $x_k$ or $\finetunedatasize$, while $\varepsilon_k$ accounts for irreducible error. 
    \item The exponent $a_k$ describes how the error scales with $x_k$ in the data-unbounded regime, where $f_k \propto x_k^{-a_k}$ as $\finetunedatasize \to \infty$. 
    \item The data scaling exponent $\datascalingexponent$ quantifies how the error decreases with increasing data size $\finetunedatasize$ for fixed inference compute, i.e., $f_k \propto \finetunedatasize^{-\datascalingexponent}$
    \item The exponent $b_k$ determines how $x_k$ affects the reducible error $\beta_k x_k^{b_k} + \xi_k$ and, thereby the impact of increasing $\finetunedatasize$. For $b_k > 0$, larger $x_k$ implies: (1) requiring more finetuning data to achieve the same error and (2) faster error reduction per additional example. See Section~\ref{sec:interpret_exponent_b_in_add_interact_parametric_form} for details.
\end{itemize}

For multiple factors $x$, we model the error as
\begin{align}
    \label{eq:add_interact_functional_form_multiple_factors}
    f(x, \finetunedatasize)
        = \sum_k \alpha_k x_k^{-a_k} + \sum_{k}\beta_k x_k^{b_k} \finetunedatasize^{-\datascalingexponent} + \xi \finetunedatasize^{-\datascalingexponent} + \varepsilon
\end{align}
where $\alpha_k, \beta_k, \xi, \varepsilon \geq 0$ and $a_k, b_k, \datascalingexponent \in \R$. Consistent with prior work \citep{alabdulmohsinGettingViTShape2023,bahriExplainingNeuralScaling2024}, we assume $\datascalingexponent$ is independent of the scaling factors.

While we primarily use \texttt{add-interact} defined in Equation~\ref{eq:add_interact_functional_form}, alternative formulations exist. These include its simplifications (e.g., \texttt{add-interact$_{s}$}),
as well as commonly used additive power-law functions (\texttt{add}) and its multiplicative variants (\texttt{mult}). Section~\ref{sec:appendix_parametric_scaling_functions} provides expressions of each parametric function we study.

\paragraph{Allocating Inference Compute}

We aim to optimize the allocation of inference compute across scaling factors $(x_N, x_T, x_V)$ to maximize model performance under a given per-example inference compute budget $\computebudget$ and finetuning data size $\finetunedatasize$. This can be formulated as follows:
\begin{align}
    \label{eq:inference_compute_optimal_scaling_opt_problem}
    x^*(\computebudget; n)
        = \argmin_{x \in \sX :\; c(x) \leq \computebudget} f(x, \finetunedatasize)
\end{align}
where $x^*$ represents the optimal scaling factors. For simplicity, we omit $\computebudget$ and $\finetunedatasize$ when their context is implicit. $x^*(\computebudget; \finetunedatasize)$ is also referred to as the ``(inference) compute-optimal frontier''.

Unlike training compute optimization, inference compute optimization does not treat $\finetunedatasize$ as an optimization variable, as finetuning data size does not affect inference compute cost. However, for certain parametric functions (e.g., \texttt{add-interact}), $\finetunedatasize$ can still influence the solution $x^*$. For others (e.g., \texttt{add}), the solution $x^*(\computebudget)$ is independent of $\finetunedatasize$. Table~\ref{tab:scaling_functional_form_expressions} summarizes which parametric functions exhibit this dependency.

For simple additive power-law parametric forms (e.g., \texttt{add} in Table~\ref{tab:scaling_functional_form_expressions}) and when inference compute is limited to LM compute costs (i.e., $c(x) = x_N x_T x_V$), the problem admits an analytic solution if $\sX$ is continuous. However, when vision model compute costs are included (e.g., $c(x)$ as defined in Equation~\ref{eq:inference_compute_of_video_vlm_with_vit}), deriving $x^*$ analytically as a function of $\computebudget$ and $\finetunedatasize$ becomes intractable. This is further complicated by the discrete nature of $\sX$ (e.g., pretrained LMs are only available in a fixed set of sizes). To address these challenges, we compute~$x^*$ using a brute-force search over all combinations in $\sX$. In practice, this is computationally feasible in our setup due to the small size of $\sX$ : $|\sX_N| \leq 5$, $|\sX_T| \leq 128$, and $|\sX_V| \leq 28$. Because the optimization problem is discrete, the optimal scaling factors $x^*$ often lie within the feasible region rather than on the boundary.

\begin{figure*}[h!]
    \centering
    \myincludegraphics[width=\textwidth]{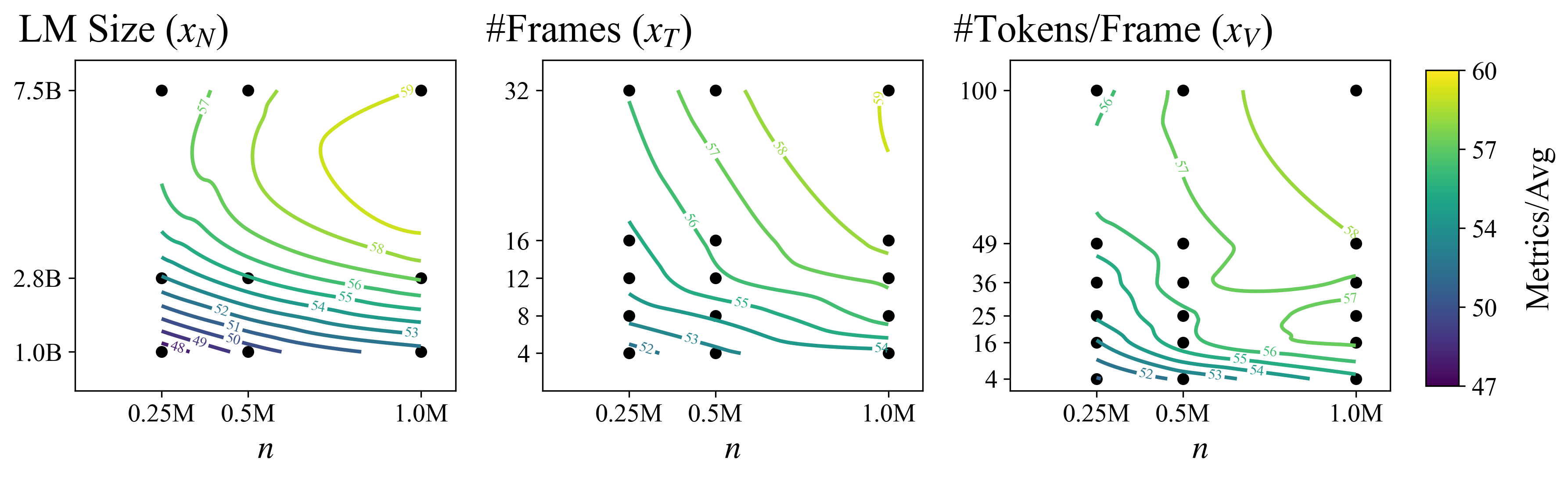}
    \caption{
        \textbf{IsoPerformance Contours.} Contours show average task performance as a function of a scaling factor (e.g., $x_N$, $x_T$, or $x_V$) and finetuning data size $\finetunedatasize$, derived from the \textit{star sweep}. As detailed in Section~\ref{sec:training_sweeps}, we construct the star sweep by starting with a inference compute-intensive ``center'' $x^{\bigstar} = (7.5\text{B}, 32, 196)$, varying one factor at a time while keeping the others fixed, and finetuning on different data sizes. For instance, in the left subfigure, each dot represents a $x_N$-parameter LM finetuned on $\finetunedatasize$ examples, with $x_T = 32$ and $x_V = 196$ fixed. Performance improves as scaling factors and $\finetunedatasize$ increase, albeit at a diminishing rate. Irregularities in the contour lines, particularly near boundaries, arise from interpolation artifacts (via \texttt{matplotlib.pyplot.contourf}) and variability in benchmark scores across fine-tuning runs.
    }
    \label{fig:plt_star_sweep_performance_wrt_shape_param_and_data_size_Metrics_Avg}
\end{figure*}

\begin{figure*}[h!]
    \centering
    \myincludegraphics[width=\textwidth]{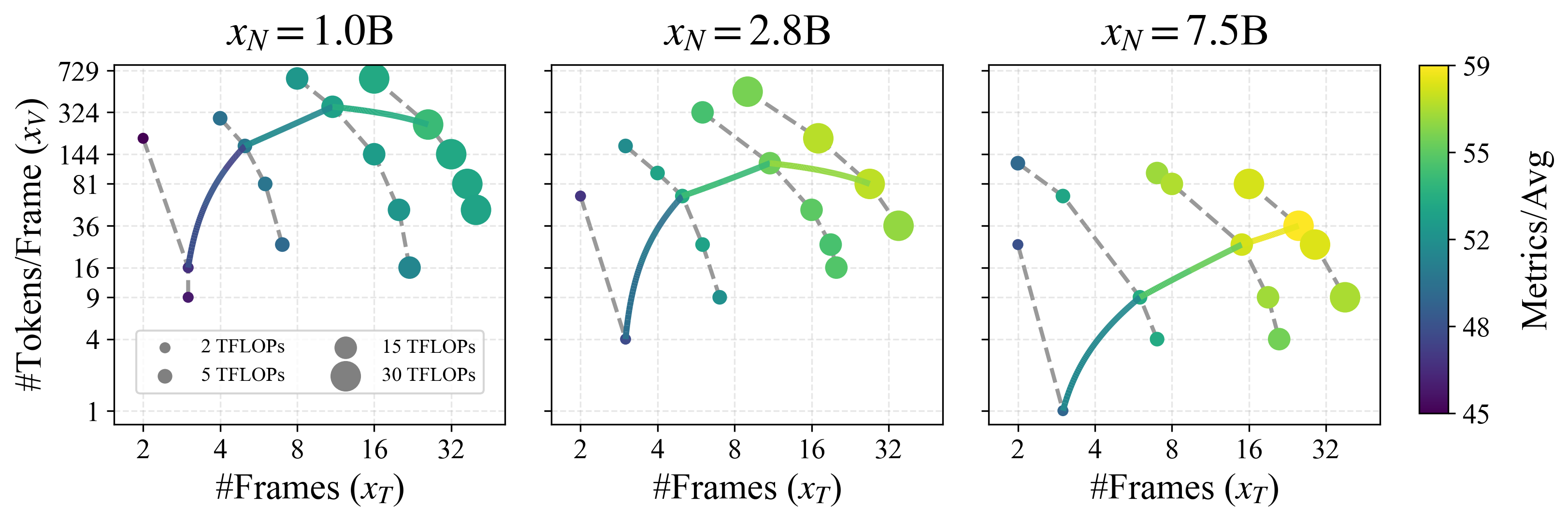}
    \caption{
    \textbf{IsoFLOP Curves and Compute-Optimal Frontier.} IsoFLOP curves (dotted lines) show task performance (color-coded) for models with fixed inference compute cost $c(x)$ across four TFLOP budgets: 2, 5, 15, and 30. The compute-optimal frontier (solid line) connects models with the best average task performance. Both are derived from the \textit{isoFLOP sweep} described in Section~\ref{sec:training_sweeps}. The compute-optimal frontier reveals that optimal performance requires scaling both $x_T$ and $x_V$ together. Moreover, at 30 TFLOPs, a model with $x_N = 7.5$B outperforms one with $x_N = 1$B, as smaller LMs cannot effectively make use of higher compute budgets (e.g., increasing from 15 to 30 TFLOPs yields minimal gain), highlighting the bottleneck imposed by LM size. These findings underscore the importance of jointly scaling $x_N$, $x_T$, and $x_V$ to maximize performance.
    }
    \label{fig:plt_isoflop_sweep_performance_trend}
\end{figure*}

\section{Implementation Details}

\paragraph{Video Instruction Dataset}

We use a comprehensive video instruction tuning dataset of $\sim 2.2$ million examples to investigate the impact of finetuning data size $\finetunedatasize$ on model performance. This dataset is compiled from a diverse range of video sources, e.g., LLaVA-Video-178K \citep{zhangVideoInstructionTuning2024}, and includes various types of instructions, e.g., chat, question-answering, and captioning. Appendix \ref{section:appendix_video_instruction_tuning_dataset} lists dataset composition and provides some explanations on how we assemble the dataset.

\paragraph{Model and Training} 

We full-parameter finetune LLaVA-like architectures \citep{liuVisualInstructionTuning2023}, which integrate a vision model, a projector, and a language model. The SoViT-400m/14 vision model \cite{alabdulmohsinGettingViTShape2023} is chosen for its performance. The projector is a two-layer MLP that converts visual features into tokens, which are then processed by the Llama-3.2 series of LMs ($x_N \in \{1\text{B}, 3\text{B}, 8\text{B}\}$) \citep{grattafioriLlama3Herd2024}.

The finetuning process involves two stages: pretraining the projector on the LCS dataset \citep{liuVisualInstructionTuning2023} and subsequently finetuning the entire model on video instruction datasets. We use the codebase and default hyperparameters from \citet{liLLaVAOneVisionEasyVisual2024}. Appendix~\ref{section:appendix_model_and_training_details} provides additional details.

Video frames are uniformly sampled at a minimum rate of 1 frame per second (fps), repeating frames if the video duration is too short to meet this rate. To downsample a 2D grid of visual features $x_W$ into a smaller grid $x_V$, we apply bilinear interpolation \citep{liLLaVAOneVisionEasyVisual2024}, implying values of $x_V$ is constrained to be perfect squares.

\paragraph{Evaluation Tasks and Metrics}

To thoroughly assess the capabilities of video VLMs, we use a diverse set of eight video tasks, including various question-answering (QA) types, e.g., captioning, open-ended, and multiple-choice questions. Evaluations include Video Detailed Caption (VDC \citealp{zhangLMMsEvalRealityCheck2024}) for detailed video descriptions, ActivityNet-QA (AQA \citealp{yuActivityNetQADatasetUnderstanding2019}) for action-related QAs, VCGBench (VCG \citealp{maazVideoChatGPTDetailedVideo2024}) for video chat capabilities, LongVideoBench (LVB \citealp{wuLongVideoBenchBenchmarkLongcontext2024}) for long video understanding, and PerceptionTest (PT \citealp{patrauceanPerceptionTestDiagnostic2023}) for fine-grained perception. Additionally, MVBench (MV \citealp{liMVBenchComprehensiveMultimodal2024a}), Video-MME (VMME \citealp{fuVideoMMEFirstEverComprehensive2024}), and Next-QA (NQA \citealp{xiaoNExTQANextPhase2021}) provide broad evaluations across diverse video tasks and domains. We measure performance using accuracy (``Acc'') for multiple-choice QA tasks and gpt-4o-mini's ratings (``Score'') for open-ended QA and captioning tasks. All metrics are standardized to a $(0, 100)$ scale to facilitate a balanced average of evaluation metrics (``Metrics/Avg''). Appendix~\ref{section:appendix_evaluation} provides additional details on evaluation.

\begin{table*}[t]
    \centering \small
        \begin{tabular}{llp{.85cm}p{.85cm}p{.85cm}p{.85cm}p{.85cm}p{.85cm}}
        \toprule
        \textbf{Form} & \textbf{Expressions for $f(x,n)$} & \multicolumn{3}{c}{\textbf{Star+IsoFLOP CV{\scriptsize (5-fold)}}} & \multicolumn{3}{c}{\textbf{Star $\rightarrow$ IsoFLOP}} \\
        &  & MSE {\scriptsize $\downarrow$} & $E_{\%}$ {\scriptsize $\downarrow$} & $R^2$ {\scriptsize $\uparrow$} & MSE {\scriptsize $\downarrow$} & $E_{\%}$ {\scriptsize $\downarrow$} & $R^2$ {\scriptsize $\uparrow$} \\
        \midrule
        \texttt{mult} & $\alpha (\prod_{k=1}^K x_k^{-a_k}) \finetunedatasize^{-\datascalingexponent} + \epsilon$ & 1.21 & 1.62 & 0.88 & 6.73 & 3.55 & \cellcolor[rgb]{0.9019607843137255,0.9529411764705882,1.0}0.45 \\
        \texttt{add} & $\sum_k \alpha_k x_k^{-a_k} + \xi \finetunedatasize^{-\datascalingexponent} + \epsilon$ & \cellcolor[rgb]{0.9019607843137255,0.9529411764705882,1.0}0.56 & \cellcolor[rgb]{0.9019607843137255,0.9529411764705882,1.0}1.11 & \cellcolor[rgb]{0.9019607843137255,0.9529411764705882,1.0}0.94 & \cellcolor[rgb]{0.9019607843137255,0.9529411764705882,1.0}2.04 & \cellcolor[rgb]{0.9019607843137255,0.9529411764705882,1.0}2.15 & \cellcolor[rgb]{0.6,0.8392156862745098,1.0}0.83 \\
        \texttt{add-interact$_{s}$} & $\sum_k \alpha_k x_k^{-a_k} + \sum_k \beta_k x_k^{b_k} \finetunedatasize^{-\datascalingexponent} + \epsilon$ & \cellcolor[rgb]{0.6,0.8392156862745098,1.0}0.24 & \cellcolor[rgb]{0.6,0.8392156862745098,1.0}0.8 & \cellcolor[rgb]{0.6,0.8392156862745098,1.0}0.97 & \cellcolor[rgb]{0.30196078431372547,0.7215686274509804,1.0}0.94 & \cellcolor[rgb]{0.30196078431372547,0.7215686274509804,1.0}1.32 & \cellcolor[rgb]{0.30196078431372547,0.7215686274509804,1.0}0.92 \\
        \texttt{add-interact} & $\sum_k \alpha_k x_k^{-a_k} + \sum_k \beta_k x_k^{b_k} \finetunedatasize^{-\datascalingexponent} + \xi \finetunedatasize^{-\datascalingexponent} + \epsilon$ & \cellcolor[rgb]{0.30196078431372547,0.7215686274509804,1.0}0.2 & \cellcolor[rgb]{0.30196078431372547,0.7215686274509804,1.0}0.77 & \cellcolor[rgb]{0.30196078431372547,0.7215686274509804,1.0}0.98 & \cellcolor[rgb]{0.6,0.8392156862745098,1.0}0.95 & \cellcolor[rgb]{0.6,0.8392156862745098,1.0}1.33 & \cellcolor[rgb]{0.30196078431372547,0.7215686274509804,1.0}0.92 \\
        \bottomrule
        \end{tabular}
    \vspace{-2pt}
    \caption{
        \textbf{Comparison of Parametric Models of Task Performance.} Evaluation of different parametric functions for modeling the average task performance under two setups: (1) \textit{in-distribution} performance using 5-fold cross-validation (CV) on combined star and isoFLOP data, and (2) \textit{extrapolation} on isoFLOP data after training on star data. Metrics include mean squared error (MSE), average relative error ($E_{\%}$), and coefficient of determination ($R^2$). The \texttt{add-interact} form, which incorporates additive power laws with interaction terms, achieves the best performance, followed by its simpler variants \texttt{add-interact$_{s}$}. These functional forms outperform simpler additive (\texttt{add}) and multiplicative (\texttt{mult}) power law models, emphasizing the importance of appropriately modeling the interactions between scaling factors $x$ and finetuning data size $n$.
    }
    \label{tab:ablate_scaling_functional_form}
\end{table*}

\section{Experiments}

\subsection{Training Sweeps}
\label{sec:training_sweeps}

To study the behavior of video VLM w.r.t. scaling factors $x$ and finetuning data size $\finetunedatasize$, we perform two types of training sweeps:
\begin{itemize}[noitemsep, left=0pt]
    \item \textit{Star sweep}: We start with an inference compute-intensive ``star center'' $(7.5\text{B}, 32, 196)$, vary one factor at a time while keeping the others constant, and finetune the model on three different data sizes $\finetunedatasize$ (in millions): 0.25, 0.5, and 1. The star sweep avoids an expensive grid search over $(x_N, x_T, x_V, \finetunedatasize)$ and instead focuses on characterizing how each scaling factor scales with $\finetunedatasize$, leading to a more accurate estimate of scaling exponents (as shown in \citealp{alabdulmohsinGettingViTShape2023}). Appendix~\ref{section:appendix_training_sweeps} provides more details on how our implementation differs slightly from the original.
    \item \textit{IsoFLOP sweep}: We adjust the scaling factors $x = (x_N, x_T, x_V)$ to maintain a fixed inference compute cost $c(x)$ across four target TFLOPs: 2, 5, 15, and 30, and finetune the model on $\finetunedatasize=2$ million examples. The isoFLOP sweep is designed to identify the optimal scaling factors for a given inference FLOP budget and to evaluate the effects of jointly scaling multiple factors. Additionally, it serves as a held-out set for model validation and selection.
\end{itemize}
Appendix~\ref{section:appendix_training_sweeps} provides a detailed description of the experimental setup and execution of these sweeps.

\importantstatement{Scaling factors $x$ and finetuning data size $\finetunedatasize$ yields diminishing return on performance.}
Figure~\ref{fig:plt_star_sweep_performance_wrt_shape_param_and_data_size_Metrics_Avg} illustrates that average task performance improves with increasing scaling factors $x$ and data size $\finetunedatasize$. However, the rate of improvement (1) diminishes as $x$ and $\finetunedatasize$ grows larger (2) varies across $x,\finetunedatasize$. These trends suggest that task performance can be effectively modeled using power-law relationships with factor-specific exponents.

\importantstatement{Jointly scale $(x_N,x_T,x_V)$ is crucial.}
Figure~\ref{fig:plt_isoflop_sweep_performance_trend} presents the isoFLOP curves, which represent models with varying $x$ and fixed inference compute costs, and the compute-optimal frontier that connects the best-performing models across different inference FLOPs budget. The compute-optimal frontier underscore the importance of increasing $x_N$, $x_T$, and $x_V$ in tandem to maximize performance. For instance, doubling the inference compute from 15 to 30 TFLOPs (by varying $x_T$ and $x_V$) yields marginal performance gains for a smaller $x_N = 1$B LM, while the same increase in inference compute provides significant improvement for a larger $x_N = 7.5$B LM, highlighting the bottleneck imposed by LM size. Similar bottleneck effects are observed for $x_T$ and $x_V$ respectively. The per-task compute-optimal frontier in Figure~\ref{fig:plt_isoflop_sweep_performance_trend_metrics=all} reaffirms the importance of jointly scaling $x$ for each task. However, this trend is not perfectly consistent due to limited robustness of the fine-tuning and evaluation process, as well as the coarse granularity of the isoFLOP sweep conducted under computational constraints.

\importantstatement{
    Utility of $(x_N,x_T,x_V)$ vary by task $\rightarrow$ so does optimal allocation of inference compute.
}
Figure~\ref{fig:plt_star_sweep_performance_wrt_shape_param_and_data_size_logscale=both_metrics=all} demonstrates that the utility of scaling factors $x$ differs significantly across downstream tasks. For example, increasing the number of frames $x_T$ yields greater benefits than scaling $\finetunedatasize$ for long video understanding tasks (e.g., LongVideoBench). In contrast, $x_T$ offers minimal gains compared to $\finetunedatasize$ for fine-grained perception tasks (e.g., PerceptionTest). These variations in utility shape the compute-optimal frontier, as shown in Figure~\ref{fig:plt_star_sweep_performance_wrt_shape_param_and_data_size_logscale=both_metrics=all}. For example, in PerceptionTest, the frontier prioritizes $x_V$ (with higher marginal benefits) over $x_T$ (with lower marginal benefits). This highlights that optimal inference compute allocation strategies should adapt to the task of interest, focusing on factors that yield the highest marginal returns on performance.

\begin{figure*}[ht]
    \centering
    \raisebox{7.5pt}{\myincludegraphics[width=.3\textwidth]{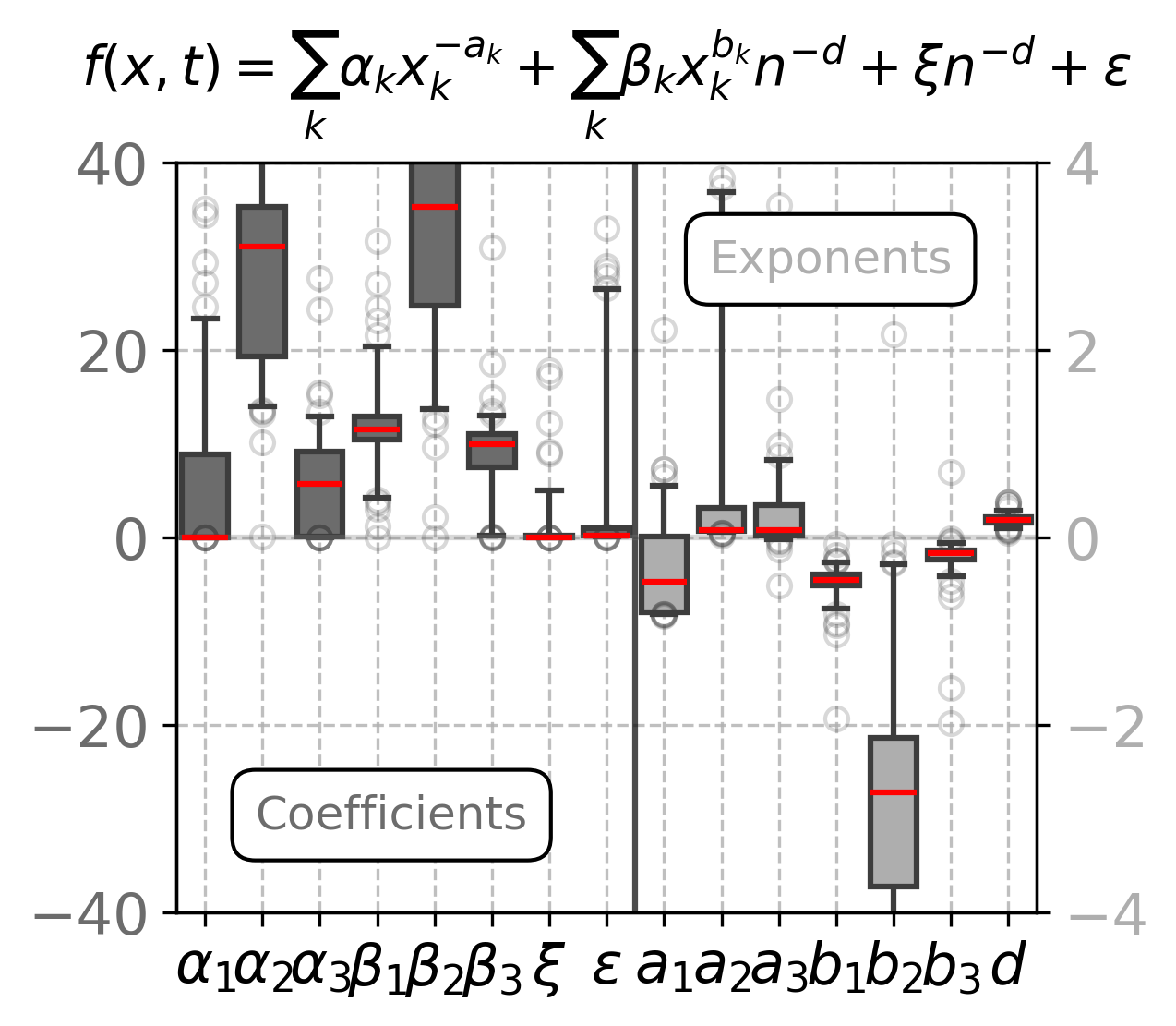}}
    \myincludegraphics[width=.33\textwidth]{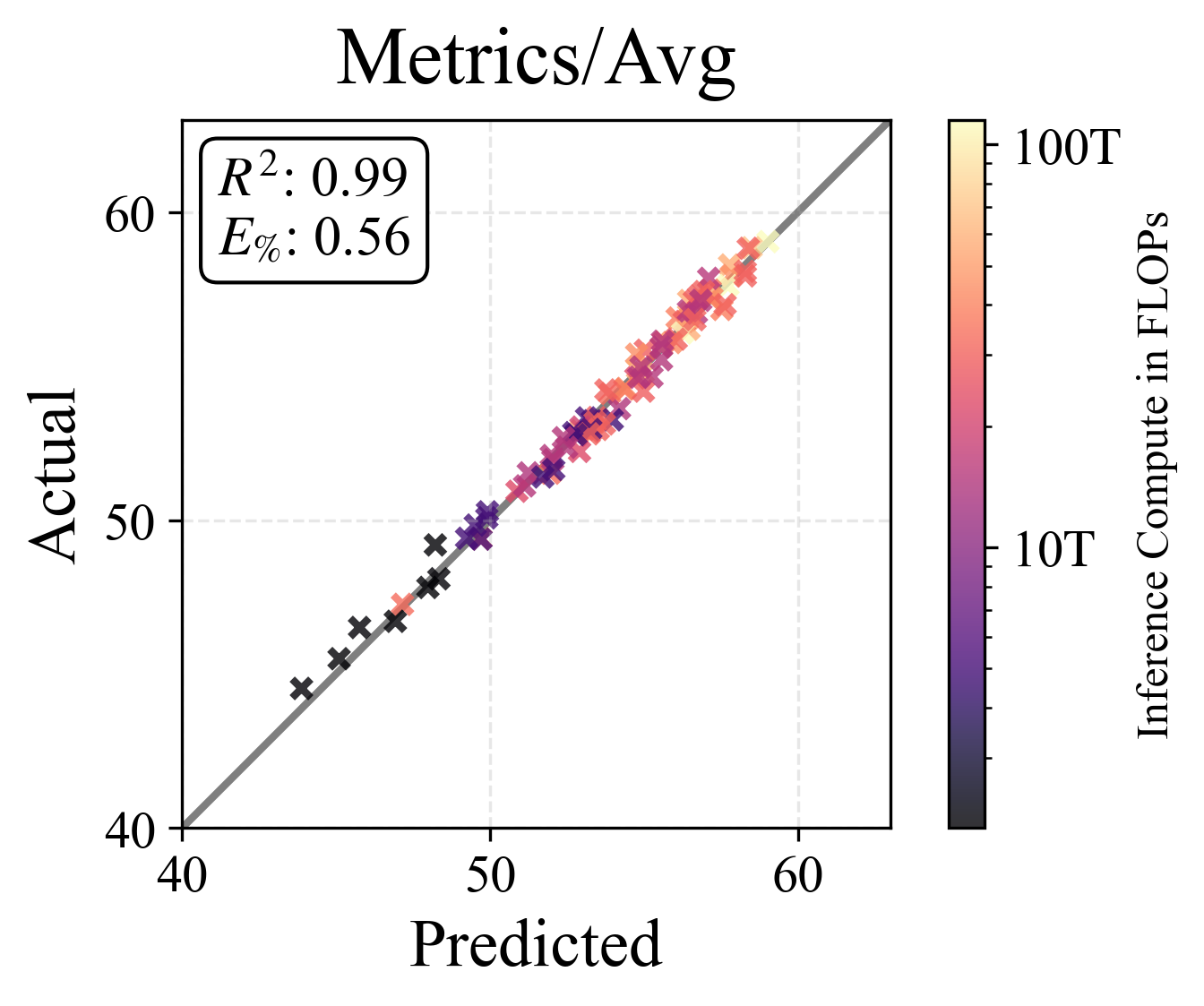}
    \raisebox{-5pt}{\myincludegraphics[width=.33\textwidth]{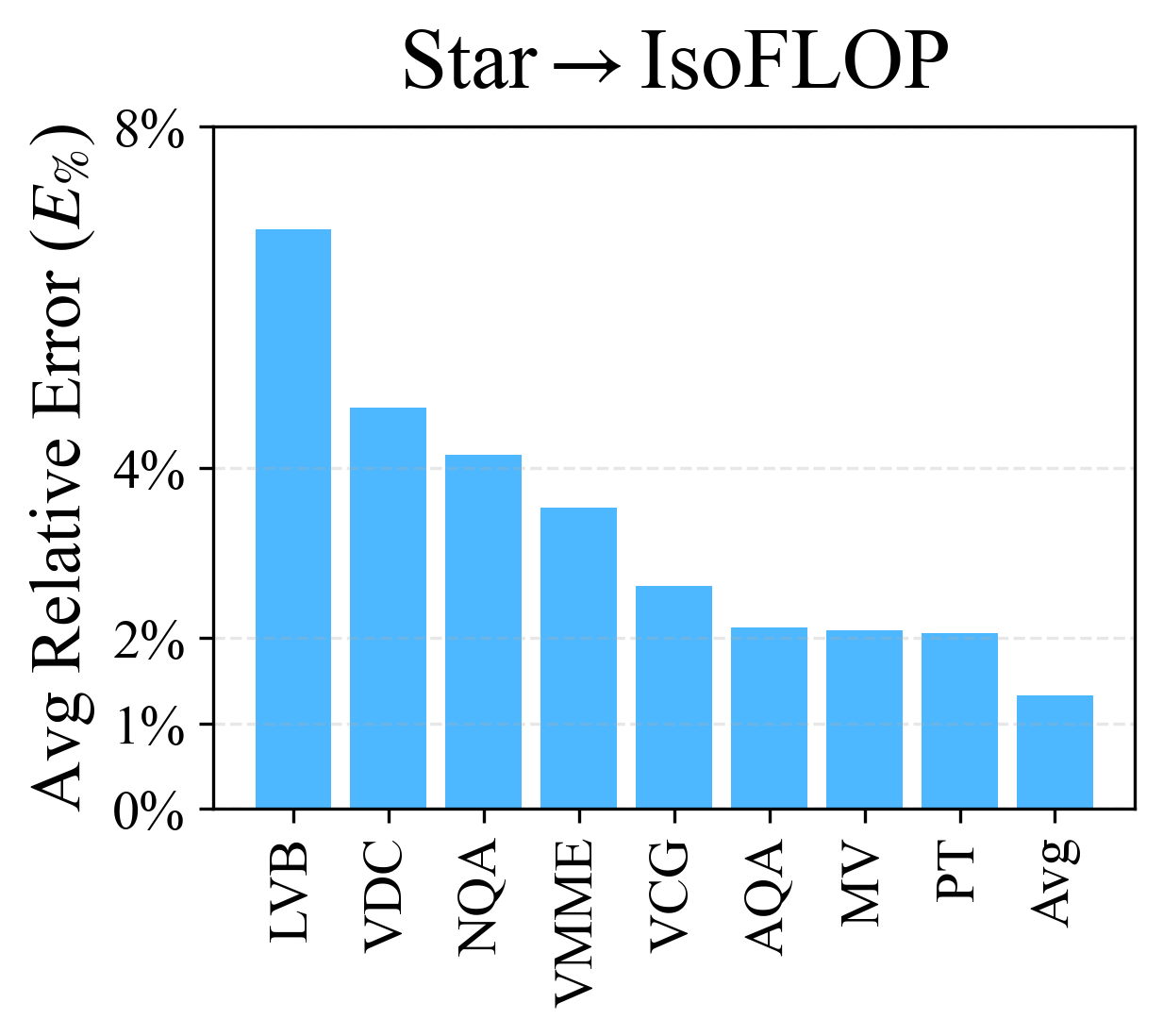}}
    
    \caption{
        \textbf{Parametric Fitting of Task Performance.} (Left) Box plot of bootstrap-resampled parameter estimates (100 resamples) for the \texttt{add-interact} model (defined in Equation~\ref{eq:add_interact_functional_form_multiple_factors}) highlights the challenge with fitting a model with just $\sim$100 examples.  (Center) Scatter plot comparing the predicted average task performance (``Metrics/Avg'') with the actual performance for each run in the star and isoFLOP sweeps. \texttt{add-interact} achieves a strong fit to data. (Right) Bar plot illustrating \texttt{add-interact}'s extrapolation performance on isoFLOP data after being trained on star data across various video tasks. While it achieves good performance for Metrics/Avg (Avg), it struggles to extrapolate effectively for tasks such as LongVideoBench (LVB) and Next-QA (NQA). 
    }
    \label{fig:parametric_fitting_of_task_performance_several_aspect_summary}
\end{figure*}

\subsection{Modeling \& Fitting of Task Performance}
\label{sec:results_modeling_and_fitting_of_task_performance}

Figure~\ref{fig:plt_isoflop_sweep_performance_trend} illustrates that the training runs sparsely cover the space of $(x,\finetunedatasize)$, making accurate interpolation or extrapolation of the compute-optimal frontier challenging. To address this, we model task performance using simple parametric functions and fit them to the training results from the star and isoFLOP sweeps. This modeling step is crucial for solving the optimization problem in Equation~\ref{eq:inference_compute_optimal_scaling_opt_problem}.

We fit the parametric model by minimizing the mean squared error loss between the predicted and observed performance metrics in log space. Appendix~\ref{sec:parametric_model_fitting_ablations} provides a detailed ablation study of the parameter estimation procedure, demonstrating the importance of a carefully designed setup for achieving accurate task performance modeling.

We assess the quality of the fit under two scenarios: (1) \textit{in-distribution} performance using 5-fold cross-validation (CV) on the star and isoFLOP data, and (2) \textit{extrapolation} performance on the isoFLOP data after estimating parameters on the star data. We report: mean squared error or MSE, average relative error $E_\% = |\hat{f}-f|/f$ in percentage, and the coefficient of determination $R^2$.

\importantstatement{
    Bagged \texttt{add-interact} is a good fit to predict the performance of (some) downstream tasks.
}
Table~\ref{tab:ablate_scaling_functional_form} compares several parametric functions (details in Appendix~\ref{sec:appendix_parametric_scaling_functions}) for modeling average task performance. Functions that include an interaction term between each scaling factor (e.g., $x_N$, $x_T$, or $x_V$) and finetuning data size $\finetunedatasize$ (e.g., \texttt{add-interact} and \texttt{add-interact$_s$}) outperforms those without interaction terms (e.g., \texttt{add}) or with incorrect interaction model (e.g., \texttt{mult}). 

Left subfigure in Figure~\ref{fig:parametric_fitting_of_task_performance_several_aspect_summary} shows a box plot of bootstrap-resampled parameter estimates for the \texttt{add-interact} model. Some parameters exhibit high variability, reflecting the challenge of robustly fitting the model with $\sim$100 examples. To alleviate this issue, we use bootstrap aggregation (bagging) to improve stability and accuracy. Ablation studies in Appendix~\ref{sec:appendix_bootstrap_aggregation} demonstrate the effectiveness of using median aggregation over 100 base models.

Based on these results, we adopt \texttt{add-interact} with bootstrap aggregation for all subsequent analyses, as it's more stable and provides the best fit. 

The middle subfigure in Figure~\ref{fig:parametric_fitting_of_task_performance_several_aspect_summary} demonstrates that \texttt{add-interact} achieves an excellent fit to the star \& isoFLOP data, covering roughly 2 orders of magnitude in inference compute. In contrast, the right subfigure in Figure~\ref{fig:parametric_fitting_of_task_performance_several_aspect_summary} reveals considerable variability in the model's extrapolation performance across tasks. While average task performance is easier to model, \texttt{add-interact} struggles with tasks like LongVideoBench (LVB) and Next-QA (NQA). For these tasks, $E_{\%} \geq 5\%$ corresponds to an average deviation exceeding $3$ points (on a 0-100 scale), representing substantial error.

\begin{figure*}[ht]
    \centering
    \myincludegraphics[width=.8\textwidth]{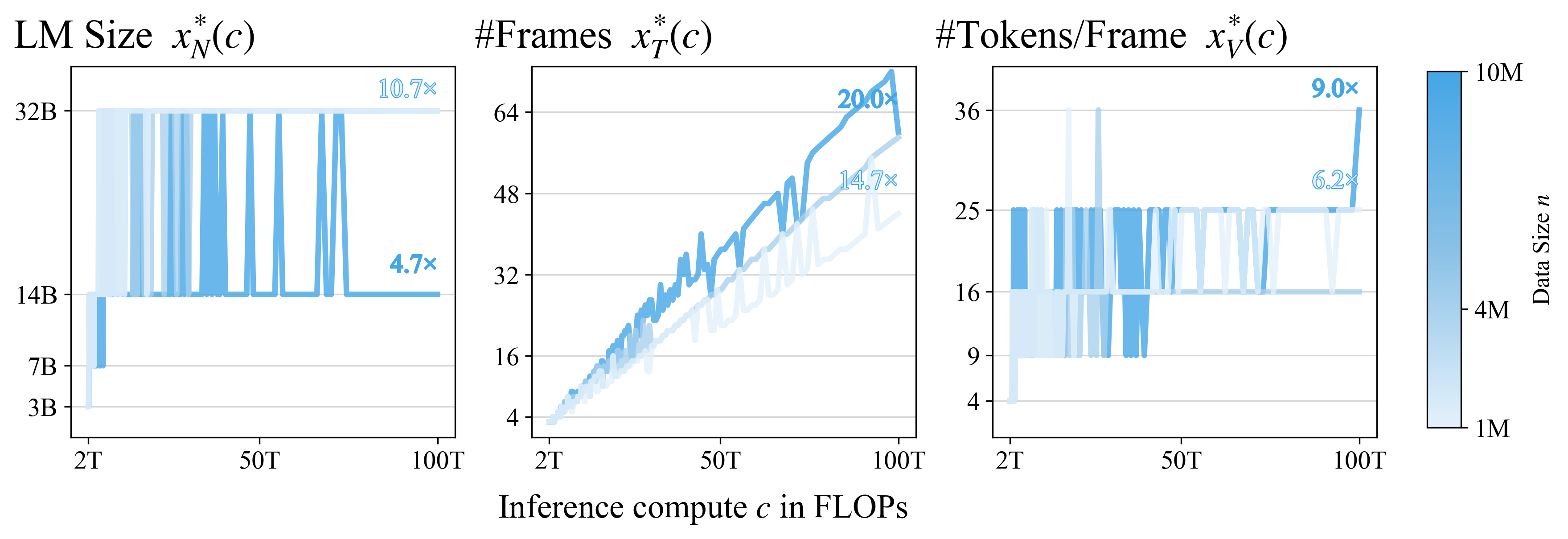}
    \raisebox{6.5pt}{\myincludegraphics[width=.18\textwidth]{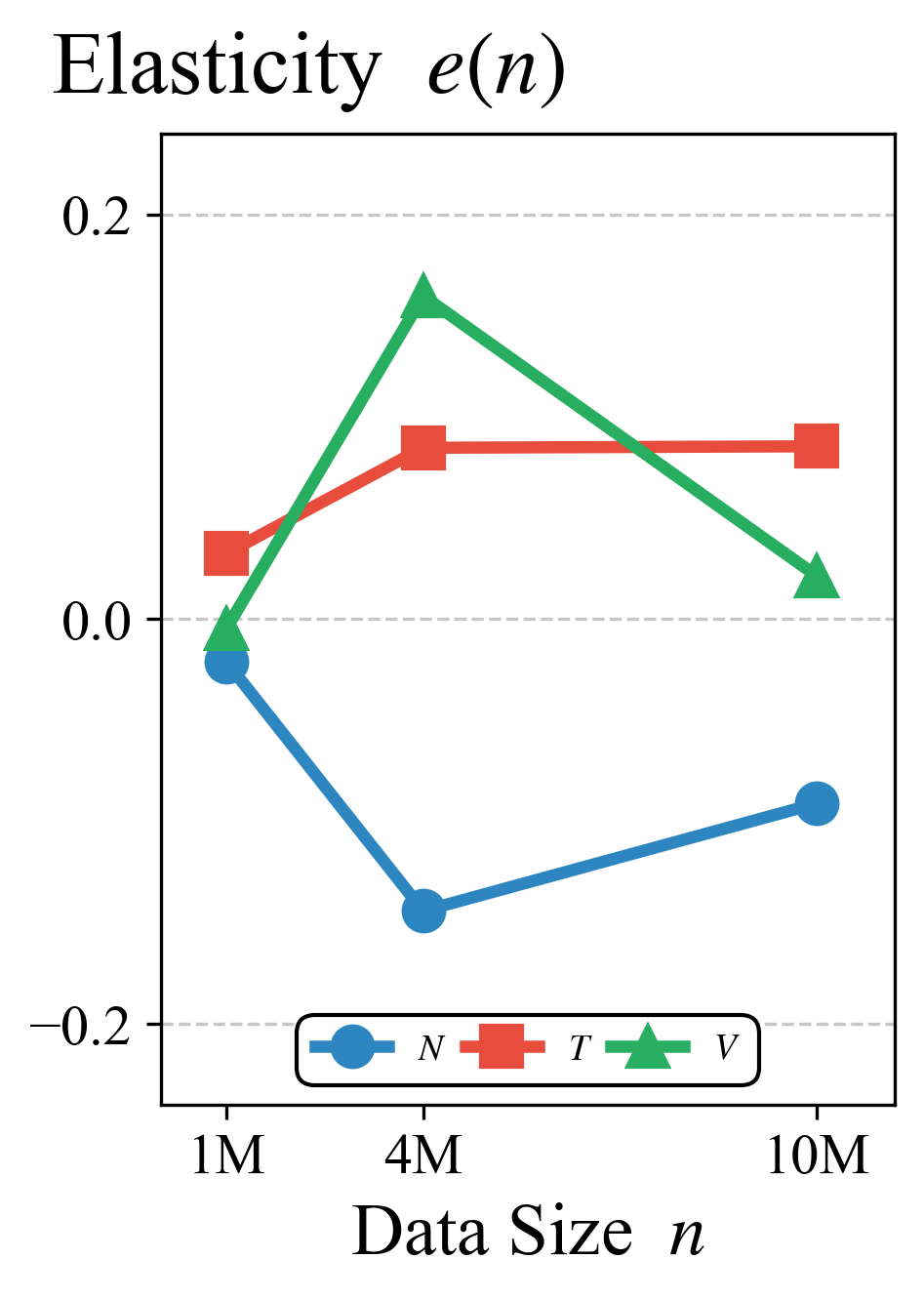}}
    
    \caption{
        \textbf{Predicted Compute-Optimal Frontier for Video VLMs.} The left three subplots show the predicted inference compute-optimal frontier $x^*(\computebudget; \finetunedatasize)$ for key scaling factors $x$ of video VLMs, across varying fine-tuning data sizes $\finetunedatasize$ (shades of blue). The blue text indicates the increase in $x^*$ as inference compute grows from $2$T to $100$T FLOPs. Task performance $f(x, \finetunedatasize)$ is modeled using the bagged \texttt{add-interact} model, which identifies an efficiency frontier requiring joint scaling of $(x_N, x_T, x_V)$ at varying rates. This frontier is non-monotonic due to the discrete domain $\sX$ of $x$. The rightmost subplot depicts the elasticity (defined in Equation~\ref{eq:elasticity_definition_as_function_of_data_size}) for each factor $k \in \{N, T, V\}$, quantifying the sensitivity of $x_k^*$ to changes in $\finetunedatasize$. For instance, as $\finetunedatasize$ increases, the frontier $x_T^*(\computebudget)$ shifts upward (in darker blue), corresponding to a positive $e_T(\finetunedatasize)$ (red curve) in the elasticity plot.
        }
    \label{fig:plt_compute_optimal_scaling_of_factors_for_fitted_scaling_function}
\end{figure*}

\subsection{Optimal Allocation of Inference Compute}

In this section, we address the question posed in Section~\ref{sec:introduction}. Using the \texttt{add-interact} model of task performance (described in Section~\ref{sec:method_parametric_model_of_performance}) that we fit in Section~\ref{sec:results_modeling_and_fitting_of_task_performance}, we compute the inference compute-optimal frontier $x^*(\computebudget; \finetunedatasize)$ for any fixed $\computebudget,\finetunedatasize$ by solving Equation~\ref{eq:inference_compute_optimal_scaling_opt_problem}. This is done via a brute-force search over $x \in \sX$ to minimize the error $f(x, \finetunedatasize)$.

\importantstatement{
    The frontier $x^*(c)$ scales jointly at varying rates and is not monotonic non-decreasing.
}
Figure~\ref{fig:plt_compute_optimal_scaling_of_factors_for_fitted_scaling_function} illustrates the predicted compute-optimal frontier $x^*(\computebudget; \finetunedatasize)$ for key scaling factors $x$ of video VLMs, across different data sizes $\finetunedatasize$. This frontier reflects the joint scaling of $(x_N, x_T, x_V)$ at different rates, consistent with empirical trends observed from training sweeps in Figure~\ref{fig:plt_isoflop_sweep_performance_trend}. The domain of $x$ is a discrete set $\sX$ with coarse increments (e.g., for $x_N$ and $x_V$), which shapes the compute-optimal frontier in two key ways: (1) it appears staggered, and (2) it is not monotonically non-decreasing, even though the predicted performance is.

Figure~\ref{fig:plt_compute_optimal_scaling_of_factors_for_fitted_scaling_function_metrics=all} shows that the predicted optimal allocation of inference compute varies notably across tasks, consistent with trends observed in Figure~\ref{fig:plt_isoflop_sweep_performance_trend_metrics=all}.

\importantstatement{
    Finetuning data size $\finetunedatasize$ influence the shape of the predicted compute-optimal frontier $x^*(\computebudget)$.
}
We quantify how changes in finetuning data size $\finetunedatasize$ affect the compute-optimal scaling factor $x_k^*(\computebudget; \finetunedatasize)$ using elasticity, defined as:
\begin{align}
    \label{eq:elasticity_definition}
    e_k(\computebudget, \finetunedatasize)
        = \frac{\partial  x_k^*(\computebudget; \finetunedatasize)}{\partial \finetunedatasize} \cdot \frac{\finetunedatasize}{x_k^*(\computebudget; \finetunedatasize)}.
\end{align}
Commonly used in economics, elasticity expresses sensitivity in percentage terms, enabling intuitive interpretation and comparison across variables of different scales. For instance, $e_T = 0.1$ indicates that a $1\%$ increase in $\finetunedatasize$ results in a $0.1\%$ increase in $x_T^*$. We use forward differences to compute elasticity and aggregate across compute budgets and data sizes to capture overall trends. Appendix~\ref{sec:appendix_details_on_elasticity_and_ablation_on_computation} provides detailed explanations of its definition, numerical approximations, and ablation studies.

\begin{figure}[ht]
    \centering
    \myincludegraphics[width=\linewidth]{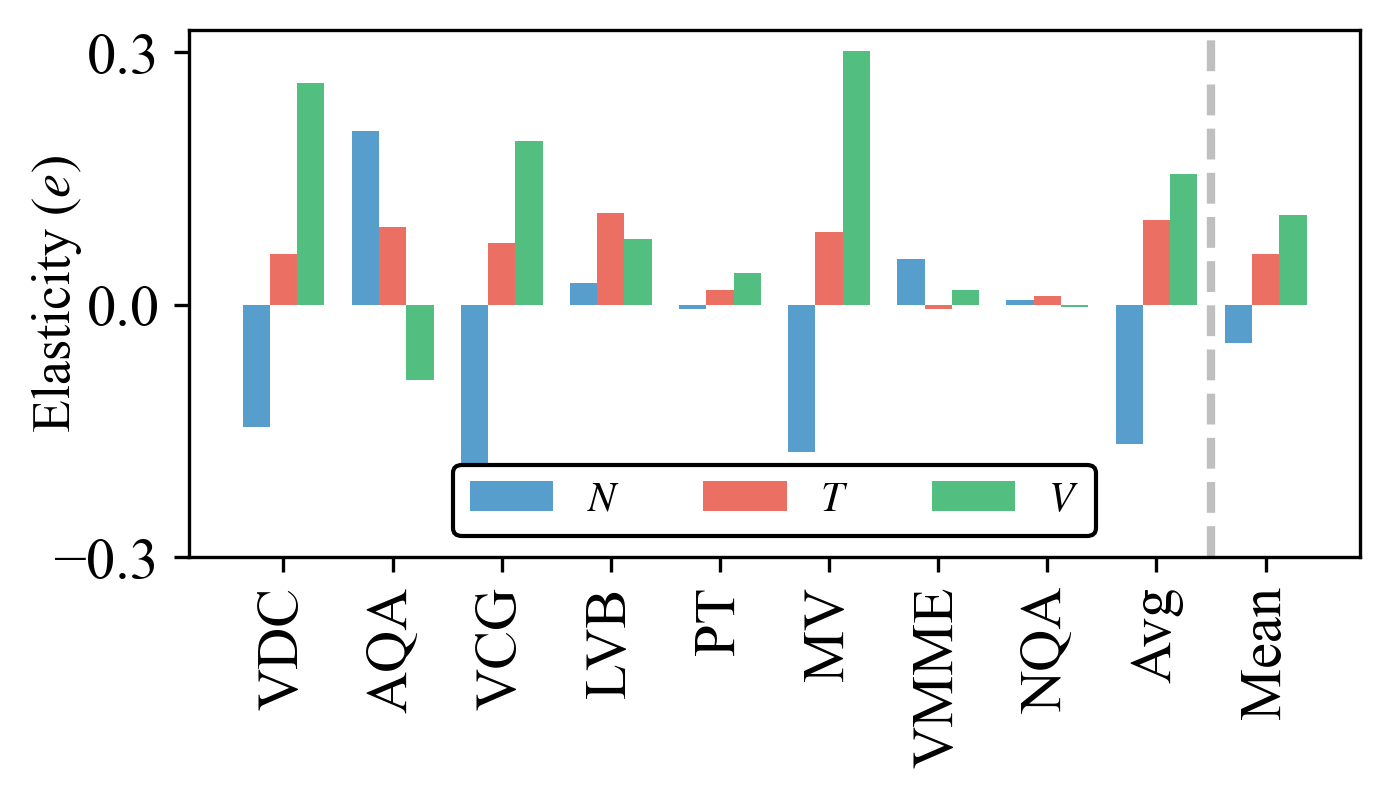}
    
    \caption{
        \textbf{Elasticity Across Tasks.} Bar plot showing the elasticity (defined in Equation~\ref{eq:elasticity_definition_as_function_of_data_size_and_compute_budget}) for scaling factors $k \in \{N, T, V\}$ across video tasks. This measures the sensitivity of optimal scaling factors $x_k^*$ to changes in data size $\finetunedatasize$. While there is significant task-specific variation, the general trend suggests decreasing $x_N$ and increasing $x_T, x_V$ as data size $\finetunedatasize$ grows.
        }
    \label{fig:plt_elasticities_across_tasks}
\end{figure}

Figure~\ref{fig:plt_compute_optimal_scaling_of_factors_for_fitted_scaling_function} illustrates the effect of $\finetunedatasize$ on the predicted compute-optimal frontier $x^*(\computebudget; \finetunedatasize)$. As $\finetunedatasize$ increases, $x_N^*$ shifts downward (negative elasticity), while $x_T^*$ and $x_V^*$ shifts upward (positive elasticity). This trend is consistent across video tasks, as shown in Figure~$\ref{fig:plt_elasticities_across_tasks}$, though with task-specific variations. The average trend suggests decreasing $x_N$ and increasing $x_T,x_V$ as data size $\finetunedatasize$ grows.

\section{Discussions}

\paragraph{Joint Scaling in Video VLMs}

Efficient use of compute through joint scaling of factors is a recurring theme in scaling law studies. EfficientNet \citep{tanEfficientNetRethinkingModel2019} demonstrated that scaling a single factor (e.g., resolution) while keeping others (e.g., width and depth) fixed is suboptimal, as bottlenecked factors limit performance gains. Instead, it introduced compound scaling, which jointly scales multiple architectural factors to achieve better results. This principle has been applied broadly, including to ViTs \citep{alabdulmohsinGettingViTShape2023} and LM pretraining \citep{hoffmannTrainingComputeoptimalLarge2022}. We empirically confirm that this principle extends to video VLMs. Specifically, jointly scaling LM size, frame count $x_T$, and visual tokens per frame $x_V$ improves performance, as shown in Figure~\ref{fig:plt_isoflop_sweep_performance_trend}. Table~\ref{tab:ablate_scaling_functional_form} further supports this, where the best-performing model, \texttt{add-interact}, identifies a compute-optimal frontier that scales all three factors together. These findings provide stronger evidence than concurrent studies, e.g., \citet{duExploringDesignSpace2024}, which assumes a parametric model without directly observing trends from training sweeps.

\paragraph{Practical Implications of Joint Scaling}

Despite the benefits of joint scaling, many existing works, e.g., MM1 \citep{mckinzieMM1MethodsAnalysis2024}, Idefics3 \citep{laurenconWhatMattersWhen2024}, and LLaVA-OneVision \citep{liLLaVAOneVisionEasyVisual2024}, rely on ablation studies that vary one factor at a time to identify good training setups. While this reduces the number of training runs, our findings suggest that, after establishing a reasonable baseline, compute resources should instead be allocated to exploring how key factors can be scaled jointly for better performance. Furthermore, many video VLMs, e.g., LLaVA-Video \citep{zhangVideoInstructionTuning2024} and Qwen-VL \citep{wangQwen2VLEnhancingVisionLanguage2024a}, fix the size of visual representation of videos while varying LM size. Our results show this approach is suboptimal for inference compute, highlighting significant opportunities for improvement.

\paragraph{Varied Scaling Rates}

Our experiments reveal that frame count $x_T$ and tokens per frame $x_V$ have distinct effects on task performance. Scaling $x_T$ yields larger performance gains than scaling $x_V$, suggesting that improving the vision model’s efficiency is more impactful than reducing the LM’s cost of processing tokens for each frame. By enabling more frames to be processed, $x_T$ scaling provides greater overall improvements. This observation also has implications for comparing models. For instance, AuroraCap \citep{chaiAuroraCapEfficientPerformant2024} compares its model to baselines with an equal number of visual tokens $x_T x_V$ per video but varies $x_V$ while fixing $x_T$ for its method and does the reverse for baselines. This inconsistency inflates the perceived performance gains of its approach.

\paragraph{Effect of Finetuning Data Size}

Our findings provide evidence on how finetuning data size affects the compute-optimal frontier. Figure~\ref{fig:plt_compute_optimal_scaling_of_factors_for_fitted_scaling_function} and~\ref{fig:plt_elasticities_across_tasks} suggest that as more data becomes available, it is optimal to allocate less compute to LM size $x_N$ and more to video visual representations $x_T$ and $x_V$. We hypothesize this is because detailed visual representations are more complex to learn and require more data to achieve comparable performance. However, as data size grows, the richer information in each visual example contributes more to overall performance, outweighing the added learning complexity. Further research is needed to validate this hypothesis.

\section{Conclusions}
In this work, we tackle the problem of optimally allocating inference compute across scaling factors that impact both compute cost and downstream task performance in video VLMs. Our approach involves: (1) conducting training sweeps to gather performance data, (2) fitting parametric models to predict task performance, and (3) solving constrained optimization problems to identify optimal trade-offs. Our work provides deeper insights into task performance trends for video VLMs, highlights effective strategies for investigating and understanding scaling behaviors, and offers practical guidance for deploying these models at scale.

\newpage

\section*{Limitations}

\paragraph{Inference Compute Estimation}
To accurately measure inference compute for video VLMs, we account for the compute cost of the vision model. However, theoretical FLOPs alone are insufficient for real-world deployments due to several overlooked factors. (1) Hardware utilization during inference often falls short of theoretical peaks, varying across implementations and hardwares. (2) Inference compute is split into two stages: prefilling, which is typically compute-bound, and decoding, which is often constrained by memory bandwidth. Our current analysis only considers the compute cost of the prefilling stage. (3) Modern efficient inference techniques, e.g., quantization \citep{dettmersGPT3int88bitMatrix2022} and speculative decoding \citep{leviathanFastInferenceTransformers2023}, add further complexity to accurately estimating inference compute.

\paragraph{Components Not Accounted For}
While we aim to include key scaling factors affecting both inference compute and task performance, some components were omitted. For instance, we did not explore scaling the size of the vision model, as most experiments were conducted using SoViT, which lacks a series of models with varying sizes. Prior work has shown that jointly scaling language and vision models is important \citep{chenScalingMultilingualVision2024}. Additionally, we hold video instruction dataset, the language model family (e.g., Llama-3.2), and downsampling methods (e.g., bilinear interpolation) fixed. While exploring the impact of these factors could provide valuable insights, we prioritized a simple, broadly applicable setup to make use the limited computational resources.

\paragraph{Parametric Model Validation}
The accuracy of our parametric models of performance is affected by experimental design choices, which may introduce implicit biases despite careful attention to implementations, training sweep design, and model fitting. For instance, we selected the Llama-3.2 model family early in the project, which lacks pretrained LM sizes between 8B and 70B. As a result, our experiments include only three LM sizes (1B, 3B, 8B), potentially limiting our ability to fully capture the relationship between LM size and task performance. While we validated the models' extrapolation performance using the isoFLOP sweep (which includes runs with lower inference compute than the star sweep), it remains uncertain how well the predicted compute-optimal frontier generalizes to significantly higher inference compute, such as 10x the current FLOPs. Addressing these issues is challenging due to the limited computational resources allocated to this project.

\bibliography{nlp_LM,multimodal,network_vision,video,misc}

\newpage
\onecolumn

\appendix

\section{Frequently Asked Questions (FAQs)}
\label{sec:appendix_faq}

We address common questions with additional detail that may be missing from the main text’s flow.

\subsection{Why finetuning compute cost is negligible compared to inference compute under heavy demand?}
\label{sec:appendix_faq_finetuning_compute_cost_is_negligible}

To compare the costs of finetuning a video vision-language model (VLM) and running inference, consider the following. The cost of finetuning scales as \(6x_N \finetunedatasize\), where \(x_N\) is the model size (number of parameters) and \(\finetunedatasize\) is the number of training tokens. In contrast, the cost of inference scales as \(2x_N \inferencedatasize\), where \(\inferencedatasize\) is the total number of tokens processed during inference. The ratio of inference cost to training cost is therefore given by:
\[
\text{Cost ratio} = \frac{2x_N \inferencedatasize}{6x_N \finetunedatasize} = \frac{\inferencedatasize}{3\finetunedatasize}.
\]
Next, assume that both training and inference costs are dominated by the tokens used to represent a video data. If each video contributes roughly the same number of tokens during training and inference, \(\finetunedatasize\) and \(\inferencedatasize\) can be treated as the number of videos instead of the number of tokens.

As an example, suppose finetuning is performed on \(\finetunedatasize = 1\) million videos. During deployment, the model processes \(\inferencedatasize = 34\) million videos per day (daily upload to TikTok in 2024). Over the course of a month (30 days), the total number of videos processed for inference becomes \(\inferencedatasize = 34 \times 30 = 1020\) million videos. Substituting these values into the cost ratio gives:
\[
\text{Cost ratio} = \frac{\inferencedatasize}{3\finetunedatasize} = \frac{1020}{3 \times 1} = 340.
\]
Thus, in this scenario, the inference cost is approximately \(340\) times higher than the finetuning cost over a one-month deployment period. This illustrates the significant computational demands of large-scale inference compared to finetuning, especially for applications involving high volumes of video data.

\begin{figure*}[ht]
    \centering
    \myincludegraphics[width=.6\textwidth]{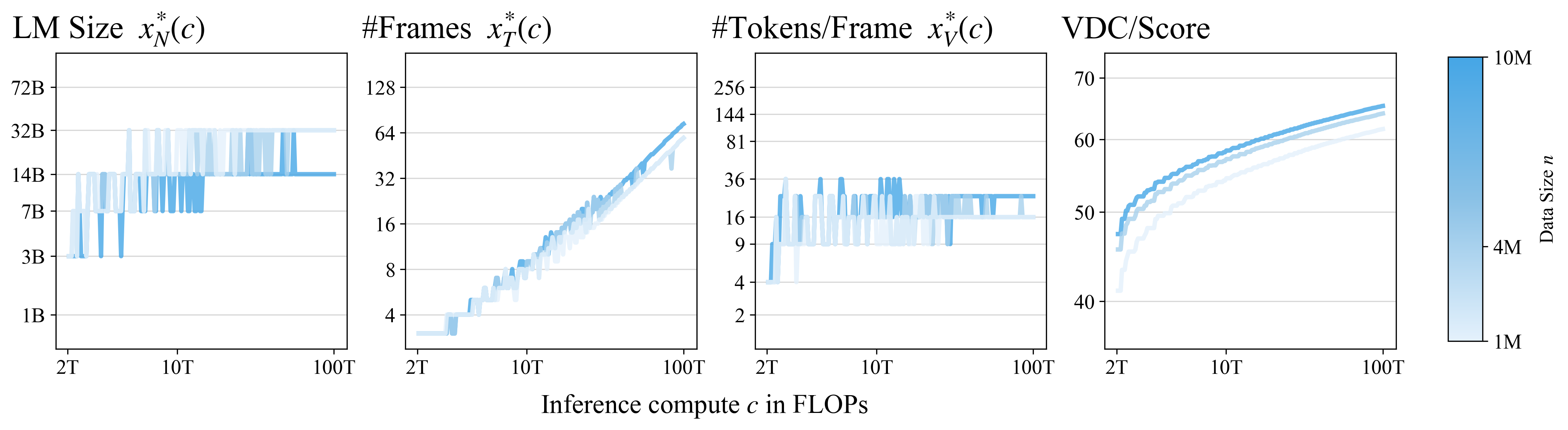}
    \myincludegraphics[width=.6\textwidth]{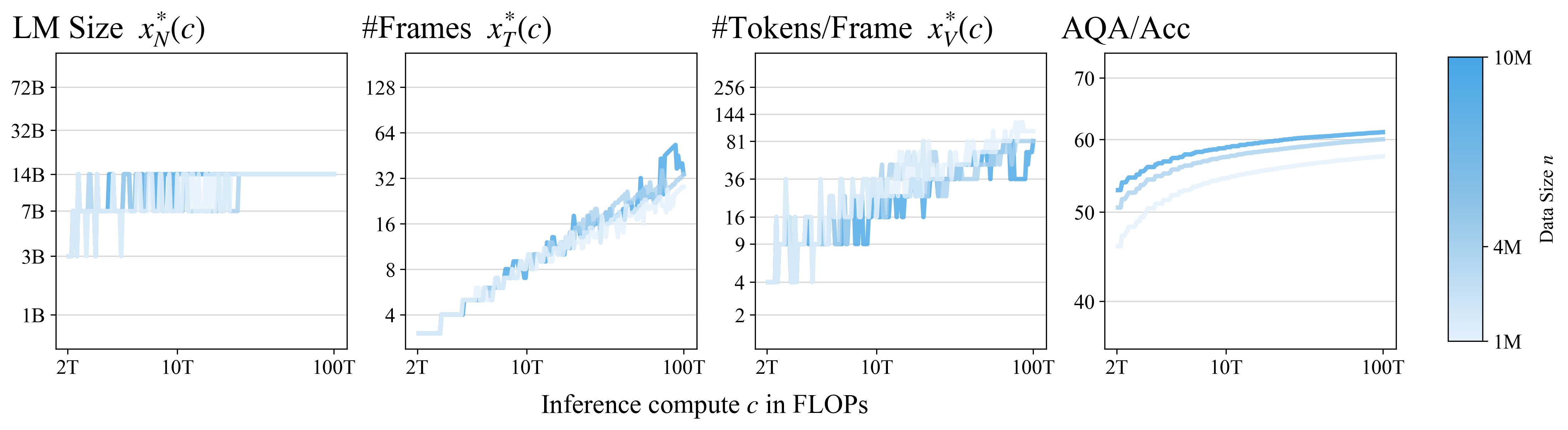}
    \myincludegraphics[width=.6\textwidth]{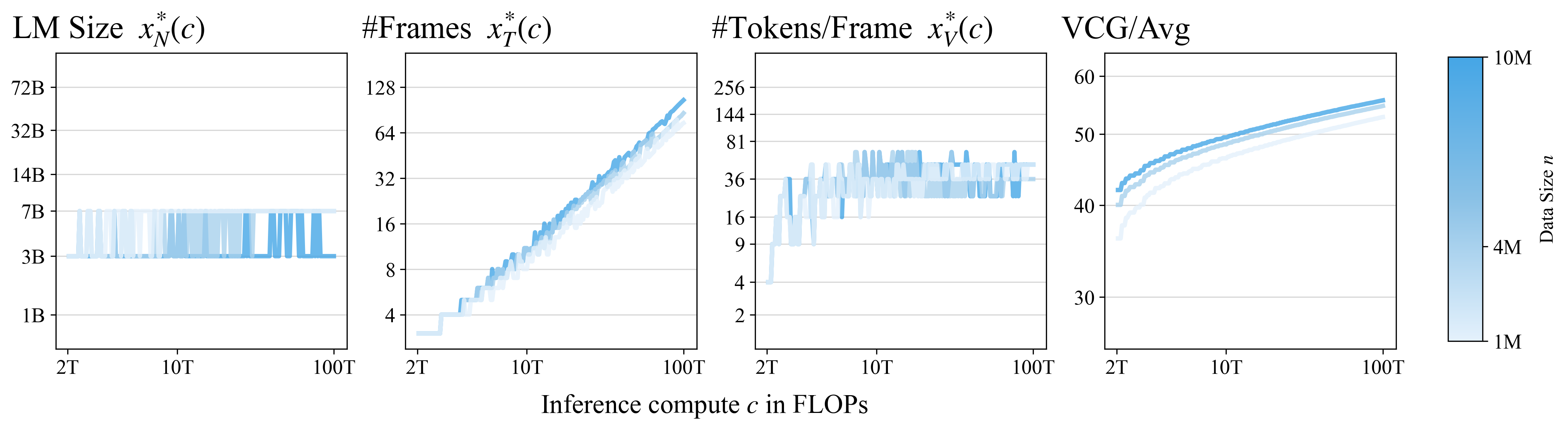}
    \myincludegraphics[width=.6\textwidth]{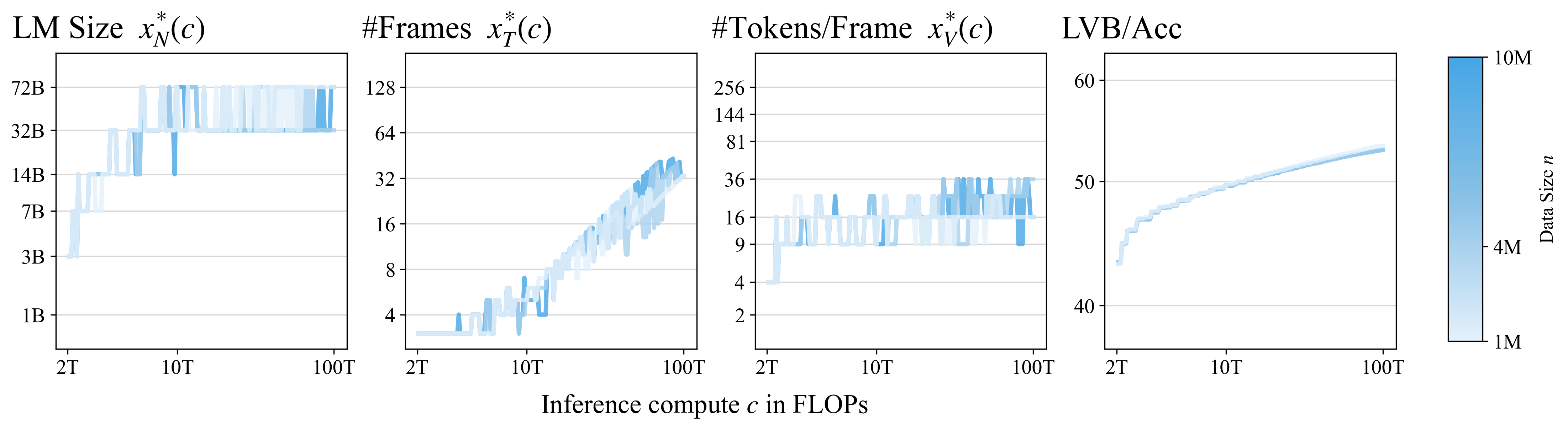}
    \myincludegraphics[width=.6\textwidth]{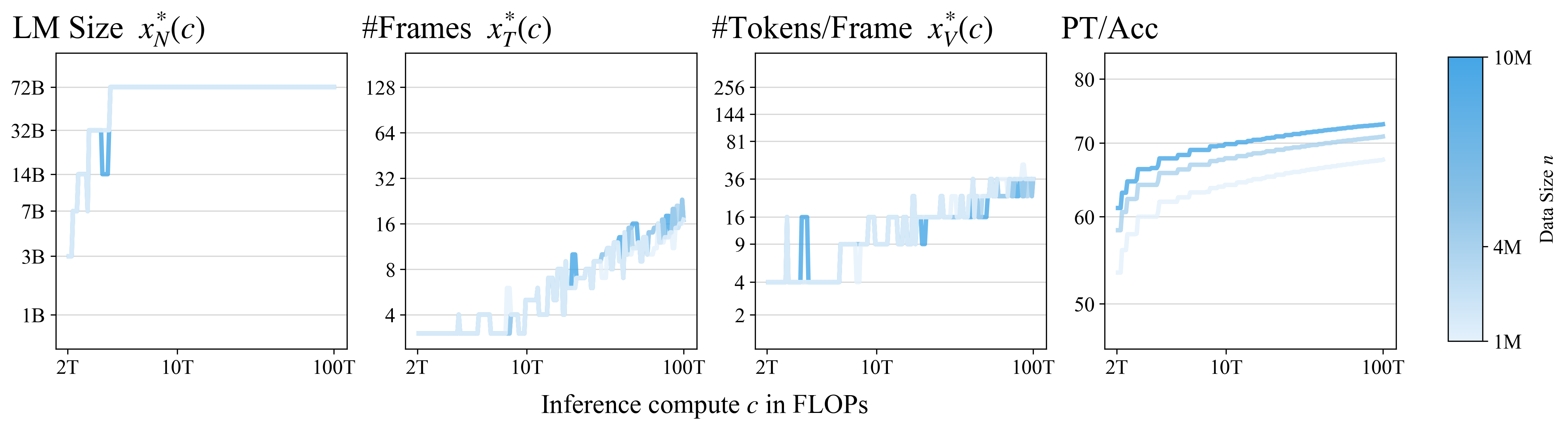}
    \myincludegraphics[width=.6\textwidth]{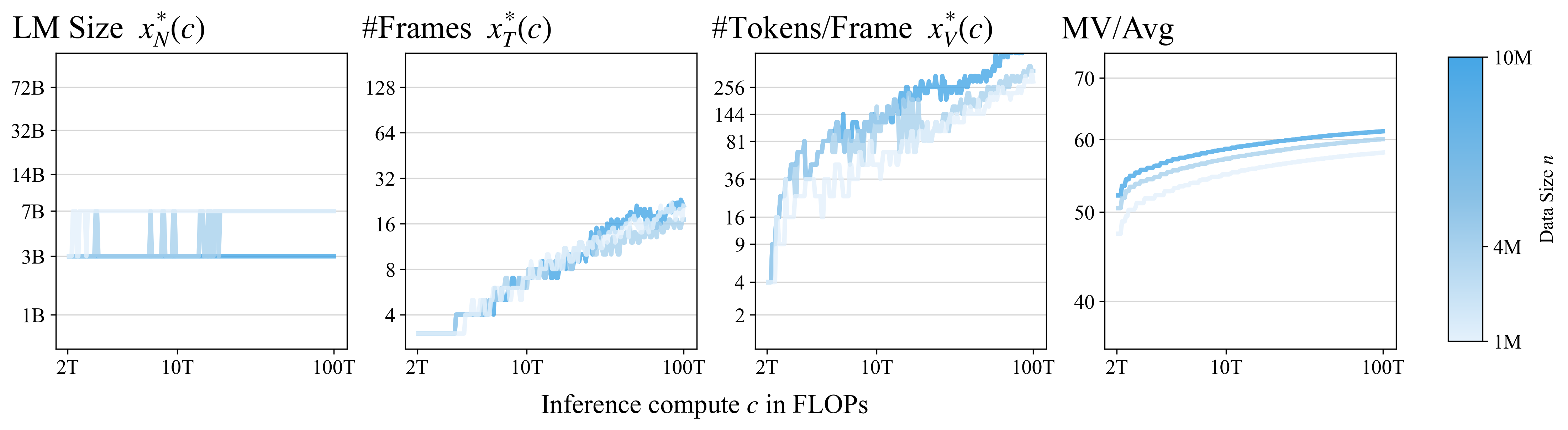}
    \myincludegraphics[width=.6\textwidth]{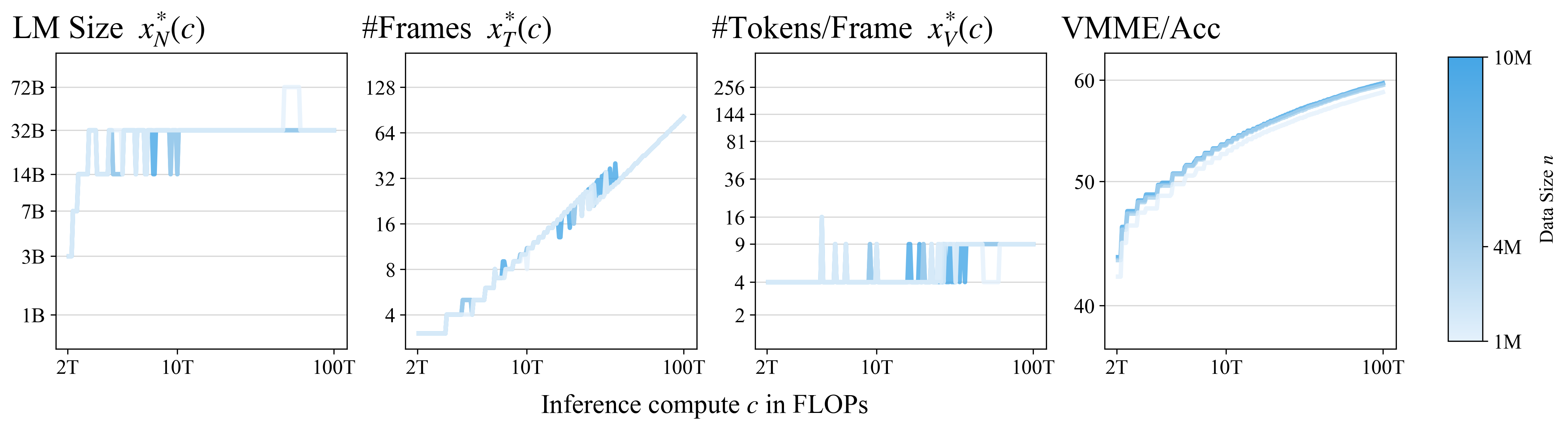}
    \myincludegraphics[width=.6\textwidth]{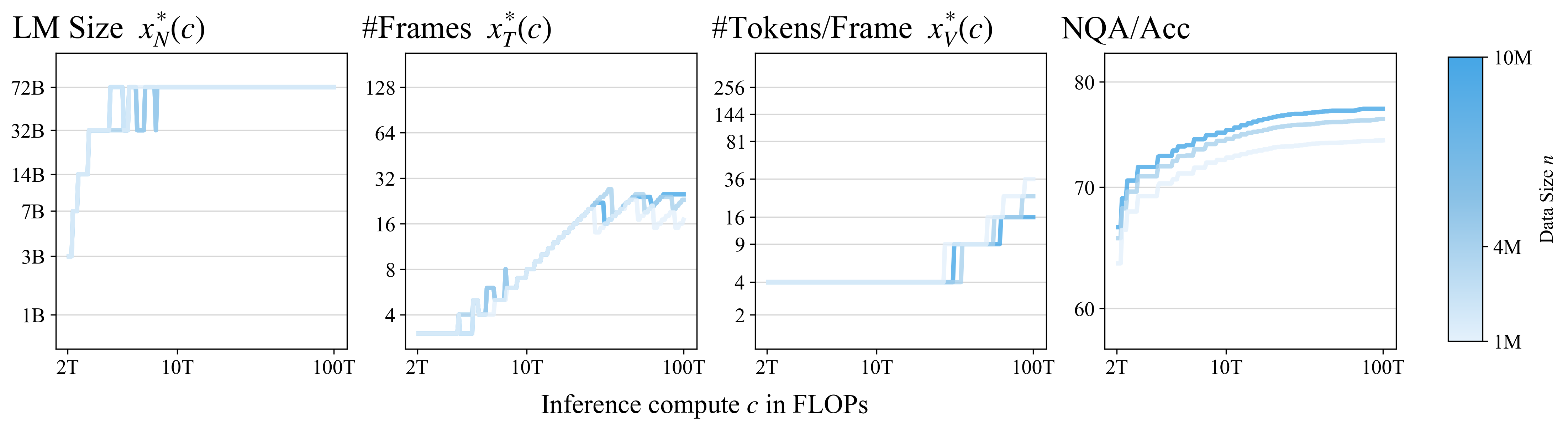}
    \caption{
      \textbf{Task-Specific Predicted Compute-Optimal Frontier.} 
      }
    \label{fig:plt_compute_optimal_scaling_of_factors_for_fitted_scaling_function_metrics=all}
  \end{figure*}

Although one might consider increasing the finetuning data size $\finetunedatasize$ to make its compute cost comparable to inference, this is rarely practical nor necessary. Collecting high-quality annotations for large-scale video data is costly and labor-intensive. Moreover, as shown in Figure~\ref{fig:plt_compute_optimal_scaling_of_factors_for_fitted_scaling_function_metrics=all}, scaling data alone yields diminishing returns, e.g., doubling the number of training videos leads to only marginal performance gains. Therefore, it is typically more effective to improve other factors, such as parameter count or number of tokens to represent a video, rather than dramatically increasing $\finetunedatasize$. This supports our assumption that the finetuning compute cost is negligible compared to inference compute in real-world deployments.

\newpage

\subsection{Why ``inference-compute optimal'' in the paper title if we perform many finetuning runs?}
\label{sec:faq_why_inferece_compute_optimal_in_title}

Our focus is on optimizing inference compute costs—the dominant cost in real-world video VLM deployments—not finetuning. As shown in Equation~\ref{eq:inference_compute_optimal_scaling_opt_problem}, our optimization constrains inference compute while treating finetuning costs as negligible. Section~\ref{sec:relationship_to_scaling_law_studies} further contrasts our work with training or finetuning compute-optimal scaling studies.

This emphasis reflects practical deployment patterns: a VLM is finetuned once or infrequently, but serves inference requests continuously at scale. As detailed in Section~\ref{sec:appendix_faq_finetuning_compute_cost_is_negligible}, monthly inference costs can far exceed the one-time finetuning cost—often by orders of magnitude.

Why, then, do we run multiple finetuning experiments? Because inference-time scaling factors (e.g., number of frames, tokens per frame) must match those used during finetuning for optimal performance. A model finetuned on 2-frame inputs performs poorly if asked to handle 64 frames at inference—it was simply never trained to process that much context. Our study jointly identifies optimal finetuning and serving configurations under a fixed inference compute budget. We believe that concurrent inference time scaling studies could also benefit from taking into account the post-training stage, e.g., if we scale inference compute by increasing the length of chain-of-thought reasoning traces, it’s beneficial to adapt the pretrained model to generate longer answers in the first place.

\newpage

\subsection{Does vision model's compute cost matter for video VLM?}
\label{sec:appendix_faq_compute_cost_of_vit_matters}

\begin{figure*}[ht]
    \centering
    \myincludegraphics[width=.6\textwidth]{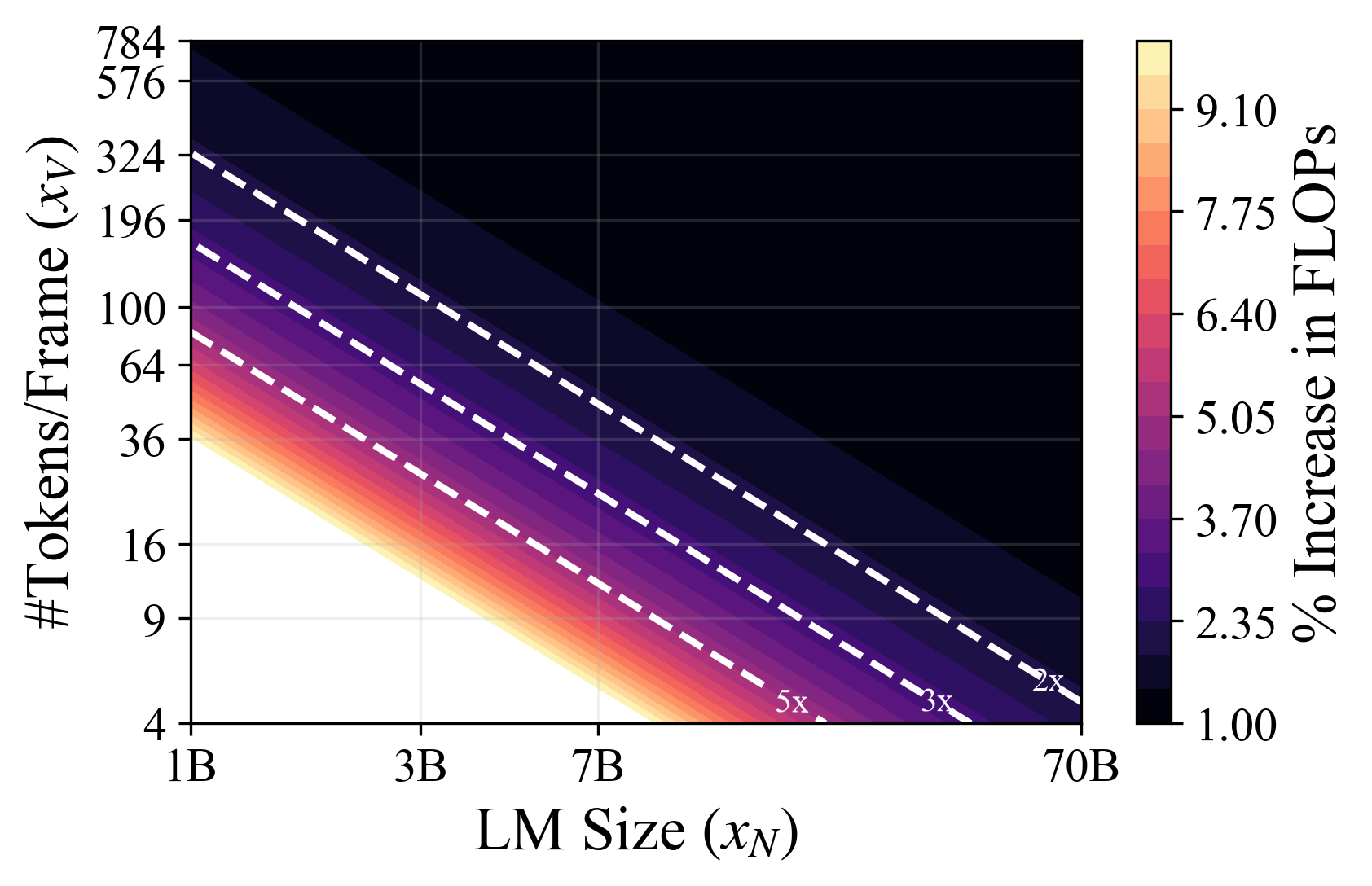}
    \caption{
        \textbf{\% FLOPs Increase From Adding Vision Model Compute.} Contours illustrates the percentage increase in inference FLOPs, when accounting for the compute cost of the vision model, as a function of language model size $x_N$ and the number of tokens per frame $x_V$. The vision encoder's compute cost significantly impacts the inference FLOPs of video VLMs. For example, for a 7B video VLM with $x_V \approx 50$, the vision model accounts for approximately half of the inference compute cost. This contribution becomes more significant for smaller $x_N,x_V$.
    }
    \label{fig:plt_inference_flops_with_and_without_vision_tower_contour}
\end{figure*}

Previous research often considers only the compute cost of the language model, neglecting the vision encoder's compute cost \citep{liInferenceOptimalVLMs2024,duExploringDesignSpace2024}. Here, we emphasize the importance of accounting for the vision model's compute cost in video VLMs.

Using the standard approximation of \(2x_N\) FLOPs per token for a transformer model with \(x_N\) parameters \citep{kaplanScalingLawsNeural2020}, the per-example inference compute costs for both the vision and language components of the video VLM are:
\[
c_{\text{ViT}} = 2x_Mx_Tx_W, \quad c_{\text{LM}} = 2x_Nx_Tx_V,
\]
where \(x_M\) is the number of vision model parameters, \(x_T\) the number of video frames, \(x_W\) the visual features output by the vision model, \(x_N\) the number of language model parameters, and \(x_V\) the visual tokens per frame after projection and resampling.

Including the vision model’s compute cost, the percentage increase in total FLOPs is:
\[
\text{\% Increase in FLOPs} = \frac{c_{\text{ViT}} + c_{\text{LM}}}{c_{\text{LM}}} = 1 + \frac{x_Mx_W}{x_Nx_V}.
\]
For a 430M-parameter SoViT-400m/14 \citep{alabdulmohsinGettingViTShape2023} outputting \(x_W = 768\) visual features, this becomes:
\[
\text{\% Increase in FLOPs} = 1 + \frac{0.43 \cdot 10^9 \cdot 768}{x_Nx_V}.
\]
The increase in compute is independent of \(x_T\), as both vision and language model costs scale linearly with the number of frames. Instead, it depends on \(x_N\) and \(x_V\), with smaller values amplifying the vision model’s relative contribution. Omitting the vision model's compute in such cases underestimates the total inference cost for video vision-language models.

Figure~\ref{fig:plt_inference_flops_with_and_without_vision_tower_contour} illustrates the percentage increase in inference FLOPs as a function of language model size $x_N$ and the number of tokens per frame $x_V$. The contours highlight the substantial impact of the vision encoder's compute cost on the overall inference FLOPs of video VLMs. Notably, for a 7B video VLM with $x_V \approx 50$, the vision model contributes to approximately half of the total inference compute cost. This influence is even more pronounced in smaller video VLMs, i.e., smaller $x_N$, where the vision model's compute cost becomes a larger fraction of the total.

\newpage
\section{Implementation Details}

\subsection{Video Instruction Tuning Dataset}
\label{section:appendix_video_instruction_tuning_dataset}

\begin{table*}[h]
    \centering
    \begin{tabular}{lrrr}
    \toprule
    \textbf{Citation} & \textbf{Data Subset} & \textbf{Size} & \textbf{Fraction (\%)} \\
    \midrule
    \cite{zhangDirectPreferenceOptimization2025} & LLaVA-Hound & 255,000 & 11.6 \\
    \cite{zhangVideoInstructionTuning2024} & LLaVA-Video-178K & 1,335,500 & 60.9 \\
    \cite{yuActivityNetQADatasetUnderstanding2019} & ActivityNet-QA & 23,530 & 1.1 \\
    \cite{xiaoNExTQANextPhase2021} & Next-QA & 34,114 & 1.6 \\
    \cite{patrauceanPerceptionTestDiagnostic2023} & PerceptionTest & 2,403 & 0.1 \\
    \cite{maazVideoChatGPTDetailedVideo2024} & VideoInstruct100K & 100,010 & 4.6 \\
    \cite{maazVideoGPTIntegratingImage2024} & VCG+112K & 112,716 & 5.1 \\
    \cite{maazVideoChatGPTDetailedVideo2024} & VideoChatGPT Human Anno. & 25,803 & 1.2 \\
    \cite{kunchangVideoChatChatCentricVideo2025} & VideoChat & 40,807 & 1.9 \\
    \cite{kayKineticsHumanAction2017} & Kinetics-710 & 39,949 & 1.8 \\
    \cite{goyalSomethingSomethingVideo2017} & Something-Something-v2 & 40,000 & 1.8 \\
    \cite{yi*CLEVRERCollisionEvents2019} & CLEVRER & 82,620 & 3.8 \\
    \cite{bainFrozenTimeJoint2021} & WebVid & 99,922 & 4.6 \\
    \midrule
    & Total & 2,192,374 & 100.0 \\
    \bottomrule
    \end{tabular}
    \caption{Video Instruction Tuning Dataset Composition}
    \label{tab:dataset_composition}
\end{table*}

We compile a comprehensive video instruction-tuning dataset designed to explore the scaling behavior of finetuning data size and to achieve a performance level that remains relevant to both researchers and practitioners. This dataset, consisting of approximately 2.2 million examples, integrates data mixture used from two prior works: VideoGPT+ \citep{maazVideoGPTIntegratingImage2024} and LLaVA-Video \citep{zhangVideoInstructionTuning2024}. 

The dataset is sourced from a diverse array of video sources, e.g., LLaVA-Video-178K, and includes various types of instructions. For example, VideoInstruct100K, VCG+112K, and VideoChat datasets contain conversational data, while other subsets primarily focus on question-answering (QA). A smaller portion is dedicated to captioning tasks. Table~\ref{tab:dataset_composition} provides additional details.

To assess the dataset's effectiveness, we conduct a coarse ablation study by removing subsets such as LLaVA-Hound, the VideoChatGPT suite, and the QA datasets. Although some tasks show improved performance in these ablated setups, the complete dataset demonstrates the best average performance across downstream tasks, underscoring its overall efficacy.

\newpage
\subsection{Model \& Training}
\label{section:appendix_model_and_training_details}

\begin{table*}[h]
    \centering
    \begin{tabular}{@{}l|cc@{}}
        \toprule
        \textbf{Hyperparameter} & \textbf{Pretraining} & \textbf{Finetuning} \\
        \midrule
        Trainable components & Projector & Vision model, projector, and language model \\
        Batch size & 512 & 256 \\
        Learning rate & 1e-3 (projector) & 1e-5 (LM \& projector), 2e-6 (vision tower) \\
        Learning rate schedule & Cosine (3\% warmup) & Cosine (3\% warmup) \\
        Weight decay & 0 & 0 \\
        Optimizer & AdamW & AdamW \\
        Epochs & 1 & 1 \\ 
        \bottomrule
    \end{tabular}
    \caption{Training hyperparameters for pretraining (the projector) and video instruction tuning.}
    \label{tab:training_hyperparams}
\end{table*}

In this paper, we explore LLaVA-like architectures \citep{liuVisualInstructionTuning2023}, which comprise three key components: a vision model, a projector, and a language model. The vision model, SoViT-400m/14 \cite{alabdulmohsinGettingViTShape2023}, is chosen for its high performance with minimal inference FLOPs and encodes each frame independently. The projector is a two-layer MLP that transforms visual features into tokens. For the language model, we use the Llama-3.2 series, available in sizes of 1B, 2.8B, and 7.5B parameters, to process these visual token representations for video-related tasks.

The finetuning process consists of two stages: pretraining the projector and video instruction tuning. Initially, we pretrain the MLP projector using the 558k LCS dataset \citep{liuVisualInstructionTuning2023}, and this pretrained weight is employed in all subsequent finetuning experiments. We then finetune the entire model on video instruction tuning datasets. 

To isolate the effects of video instruction tuning, we deliberately avoid using checkpoints that have been further fine-tuned on datasets specifically designed to enhance knowledge understanding or improve instruction following for images or multi-image inputs. In contrast, models like LLaVA-OneVision \citep{liLLaVAOneVisionEasyVisual2024} and LLaVA-Video \citep{zhangVideoInstructionTuning2024} incorporate such additional fine-tuning stages to boost performance. While this allows us to focus solely on evaluating the impact of video instruction tuning, it likely limit our model's performance compared to these approaches.

We adopt the codebase and default hyperparameters from \citet{liLLaVAOneVisionEasyVisual2024}. Table~\ref{tab:training_hyperparams} details the specific hyperparameters used during both pretraining and video instruction tuning.

\newpage
\subsection{Evaluation}
\label{section:appendix_evaluation}

\begin{table*}[h]
    \centering
    \small
    \begin{tabular}{l|llcllc}
    \toprule
    \textbf{Evaluation} & \textbf{Abbr.} & \textbf{QA Type} & \textbf{Long Vid.} & \textbf{LLM Judge} & \textbf{LLMs-Eval Task Name} & \textbf{Metrics} \\
    \midrule
    VideoDetailedDescription & VDC & caption & & gpt-4o-mini & \texttt{video\_dc499} & score \\
    ActivityNet-QA & AQA & open-ended & & gpt-4o-mini & \texttt{activitynetqa} & accuracy \\
    VCGBench & VCG & open-ended & & gpt-4o-mini & \texttt{videochatgpt} & score \\
    LongVideoBench & LVB & MC & \checkmark & & \texttt{longvideobench\_val\_v} & accuracy \\
    PerceptionTest & PT & MC & & & \texttt{perceptiontest\_val\_mc} & accuracy \\
    MVBench & MV & MC & & & \texttt{mvbench} & accuracy \\
    Video-MME & VMME & MC & \checkmark & & \texttt{videomme} & accuracy \\
    Next-QA & NQA & MC & & & \texttt{nextqa\_mc\_test} & accuracy \\
    \bottomrule
    \end{tabular}
    \caption{
        \textbf{Video Evaluation Benchmarks.} This table summarizes the evaluation benchmarks used to assess the capabilities of video VLMs. The benchmarks cover various question-answering (QA) types, including captioning, open-ended questions, and multiple-choice (MC) questions. Some benchmarks are designed for specific capabilities, e.g., long video understanding (under ``Long Vid.''), while others evaluate a whole suite of capabilities, e.g., Next-QA and Video-MME. The table also specifies whether a language model judge (e.g., gpt-4o-mini) is involved, the corresponding task name in LMMs-Eval \citep{zhangLMMsEvalRealityCheck2024}, and the metrics used for assessment.
        }
    \label{tab:details_of_video_evaluation_benchmarks}
\end{table*}

In this paper, we evaluate the video VLM on a diverse set of 8 downstream video tasks to ensure comprehensive assessment across various tasks and domains. For reproducibility, we use LMMs-Eval \citep{zhangLMMsEvalRealityCheck2024}. The evaluation includes: (1) VideoDetailedCaption (VDC \citealp{zhangLMMsEvalRealityCheck2024}) for detailed video descriptions, (2) ActivityNet-QA (AQA \citealp{yuActivityNetQADatasetUnderstanding2019}) for action-related QA, (3) VCGBench (VCG \citealp{maazVideoChatGPTDetailedVideo2024}) for assessing video chat capabilities, (4) 
LongVideoBench (LVB \citealp{wuLongVideoBenchBenchmarkLongcontext2024})
for long video understanding, (5) PerceptionTest (PT \citealp{patrauceanPerceptionTestDiagnostic2023}) for fine-grained perception evaluation, and (6) MVBench (MV \citealp{liMVBenchComprehensiveMultimodal2024a}), Video-MME (VMME \citealp{fuVideoMMEFirstEverComprehensive2024}), and Next-QA (NQA \citealp{xiaoNExTQANextPhase2021}) for broad evaluation across diverse video tasks and domains. This combination of benchmarks enables a thorough and reproducible analysis of the model's performance.

Table~\ref{tab:details_of_video_evaluation_benchmarks} offers a detailed summary of the video evaluation tasks employed to evaluate the capabilities of video VLMs. These tasks are categorized by the type of question-answering (QA) they address, including captioning, open-ended, and multiple-choice (MC) questions. The table highlights key features of each task, such as their emphasis on long video understanding (under ``Long Vid.''). It also specifies whether a language model judge, like gpt-4o-mini, is involved in the evaluation process. Prior work rely on gpt-3.5-turbo-0613 as the LLM judge that has since been deprecated, we transition to use gpt-4o-mini-2024-07-18 instead. Each task is linked to a specific task name within the LMMs-Eval framework \citep{zhangLMMsEvalRealityCheck2024}, and the metrics used for assessment, such as scores generated by the LLM judge or multiple-choice accuracy, are also provided. To ensure consistency, we convert both accuracy and the LLM judge's scores to $(0, 100)$, allowing for a balanced average across evaluation metrics. We use ``Metrics/Avg'' to denote the average task performance.

\newpage
\section{Training Sweeps}
\label{section:appendix_training_sweeps}

\begin{table*}[ht]
    \small
    \centering
    \begin{tabular}{lccccl}
    \toprule
    {\normalsize \textbf{Sweep}} & {\normalsize $\finetunedatasize$} & {\normalsize $x_N$} & {\normalsize $x_T$} & {\normalsize $x_V$} & {\normalsize \textbf{Comment}} \\
     & \textbf{(in million)} & \textbf{(in billion)} & & & \\
    \midrule
    Star & \{0.25, 0.5, 1\} & \{1, 2.8, 7.5\} & 32 & 196 & Vary $x_N$ \\
     & \{0.25, 0.5, 1\} & 7.5 & \{4, 8, 12, 16, 32\} & 196 & Vary $x_T$ \\
     & \{0.25, 0.5, 1\} & 7.5 & 32 & \{4, 16, 25, 36, 49, 100, 196\} & Vary $x_V$ \\
    \midrule
    IsoFLOP & 2 & 1 & \multicolumn{2}{c}{\{(2, 196), (3, 9), (3, 16)\}} & $c(x)\approx 2$ {\scriptsize TFLOPs} \\
     & & 2.8 & \multicolumn{2}{c}{\{(2, 64), (3, 4)\}} & \\
     & & 7.5 & \multicolumn{2}{c}{\{(2, 25), (3, 1)\}} & \\
    \cmidrule{3-6}
     & & 1 & \multicolumn{2}{c}{\{(4, 289), (5, 169), (6, 81), (7, 25)\}} & $c(x)\approx 5$ {\scriptsize TFLOPs} \\
     & & 2.8 & \multicolumn{2}{c}{\{(3, 169), (4, 100), (5, 64), (6, 25), (7, 9)\}} & \\
     & & 7.5 & \multicolumn{2}{c}{\{(2, 121), (3, 64), (6, 9), (7, 4)\}} & \\
    \cmidrule{3-6}
     & & 1 & \multicolumn{2}{c}{\{(8, 625), (11, 361), (16, 144), (20, 49), (22, 16)\}} & $c(x)\approx 15$ {\scriptsize TFLOPs} \\
     & & 2.8 & \multicolumn{2}{c}{\{(6, 324), (11, 121), (16, 49), (19, 25), (20, 16)\}} & \\
     & & 7.5 & \multicolumn{2}{c}{\{(7, 100), (8, 81), (15, 25), (19, 9), (21, 4)\}} & \\
    \cmidrule{3-6}
     & & 1 & \multicolumn{2}{c}{\{(16, 625), (26, 256), (32, 144), (37, 81), (40, 49)\}} & $c(x)\approx 30$ {\scriptsize TFLOPs} \\
     & & 2.8 & \multicolumn{2}{c}{\{(9, 484), (17, 196), (22, 121), (27, 81), (35, 36)\}} & \\
     & & 7.5 & \multicolumn{2}{c}{\{(16, 81), (25, 36), (29, 25), (38, 9)\}} & \\
    \bottomrule
    \end{tabular}
    \caption{
        \textbf{Star and IsoFLOP Sweep Setup}. Scaling factors $x$ and finetuning data sizes $\finetunedatasize$ used in the experiments.
      }
    \label{tab:scaling_factors_used_for_sweeps}
\end{table*}

\begin{figure*}[ht]
    \centering
    \myincludegraphics[width=\textwidth]{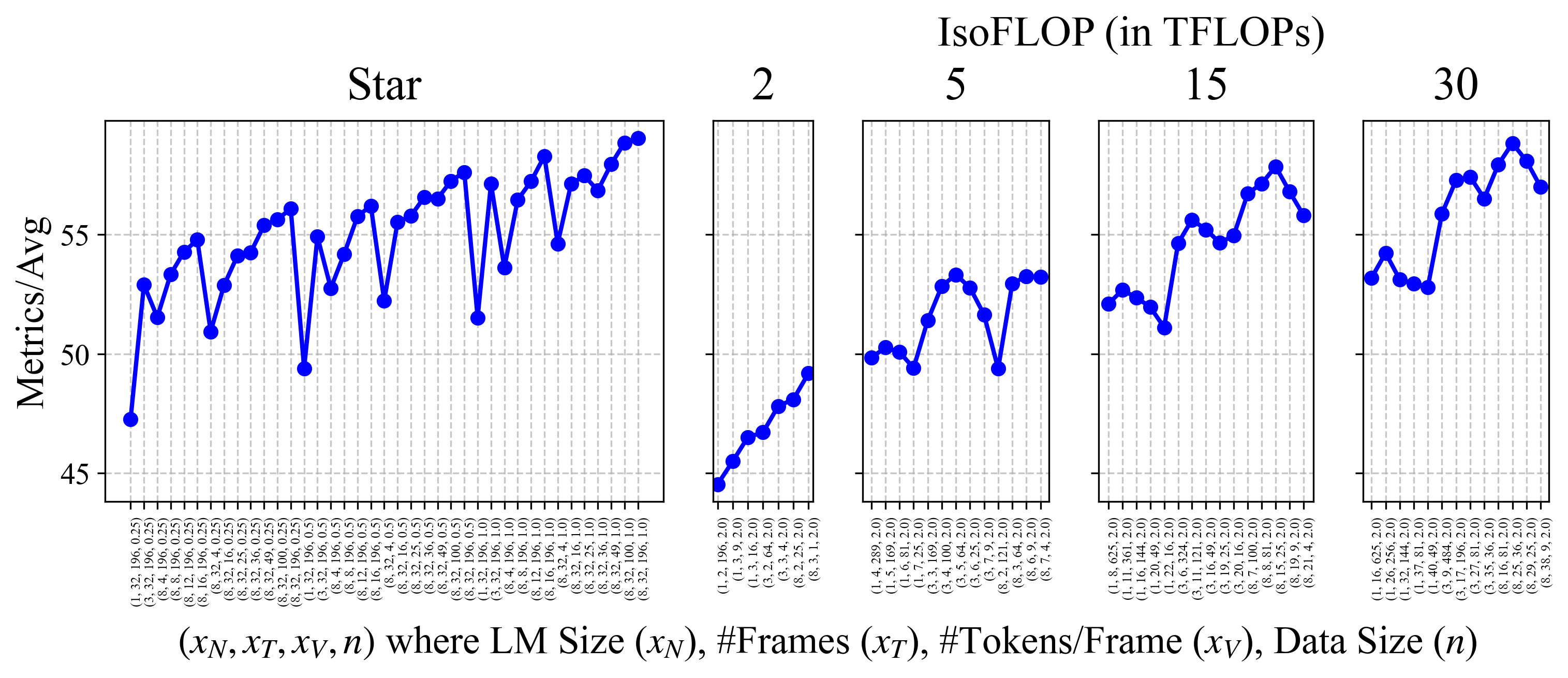}
    \caption{
      \textbf{Empirical Data from Star and IsoFLOP Sweeps}. We show the average task performance across different runs of the star and isoFLOP sweeps. The data gathered from sweeps is used for visualization and parametric fitting of scaling curves.
      }
    \label{fig:empirical_data_from_sweeps}
\end{figure*}

In this paper, we conduct two types of sweeps on scaling factors $x$ and finetuning data size $\finetunedatasize$: (1) the star sweep, proposed in SoViT \citep{alabdulmohsinGettingViTShape2023}, and (2) the isoFLOP sweep, used by Chinchilla \citep{hoffmannTrainingComputeoptimalLarge2022}. We finetune a video VLM using scaling factor $x$ on datasets of size $\finetunedatasize$, and evaluate the model to obtain data points $((x,n), f(x,n))$. These data points are used for visualization and parametric fitting. Table~\ref{tab:scaling_factors_used_for_sweeps} lists the scaling factors $x$ and finetuning data sizes $\finetunedatasize$ used in our experiments. Figure~\ref{fig:empirical_data_from_sweeps} illustrates the average task performance for each run of the star and isoFLOP sweeps.

\paragraph{Star Sweep} 

We start with an inference compute-intensive set of scaling factors, the ``star center'' $x^{\bigstar} = (7.5\text{B}, 32, 196)$. We vary one factor at a time while keeping the others fixed, and finetune the video VLM on three data sizes: $\{0.25\text{M}, 0.5\text{M}, 1\text{M}\}$. This approach allows for more accurate estimation of the scaling exponent for each factor as the finetuning data size increases. Additionally, it enables exploration on how individual scaling factor interact with finetuning data size.

The star sweep approach avoids a brute-force grid search over $(x_N,x_T,x_V,\finetunedatasize)$ to estimate scaling parameters across all dimensions. In \citet{alabdulmohsinGettingViTShape2023}, the star center $x^{\bigstar}$ is set larger than the scaling factors used in the sweep to prevent bottlenecks. For example, with a star center having \texttt{MLP dim} of 6144, the sweep explores values for \texttt{MLP dim} in a grid $(1088, 1360, 1728, 2160, 2592, 3072)$, all much smaller than 6144. In contrast, our sweep matches the largest scaling factor to the star center, with the next largest about half its value, followed by more densely sampled smaller values. For instance, with $x^{\bigstar}_T = 32$, we use the grid $(4, 8, 12, 16, 32)$ for the sweep. This provides more data points where scaling factors are smaller than the star center, and fitting a parametric function with equal weighting will prioritize these smaller values. Due to computational constraints, we set a smaller star center than the ideal, e.g., $(70\text{B}, 64, 512)$, resulting in a minor deviation from the prescribed strategy, which we consider reasonable given the trade-offs.

\paragraph{IsoFLOP Sweep} 

We vary scaling factors $x = (x_N, x_T, x_V)$ to maintain a constant inference compute cost $c(x)$ (defined in Equation~\ref{eq:inference_compute_of_video_vlm_with_vit}) across four target FLOPs: 2, 5, 15, 30. Each model is finetuned on approximately 2 million examples. For each target FLOP, we exhaustively search for scaling factor combinations within $\epsilon = 0.03$ of the target and select a well-spaced subset to ensure sufficient coverage while minimizing the number of runs. Additional points are added post-sweep to ensure the empirical compute-optimal frontier for average downstream task performance lies within each isoFLOP curve, as shown in Figure~\ref{fig:empirical_data_from_sweeps}. The isoFLOP sweep addresses key questions, such as identifying the optimal set of scaling factors for a given inference FLOP budget and assessing the impact of jointly scaling multiple factors. It also serves as a held-out validation set for model evaluation and selection.

\newpage
\section{Modeling and Fitting of Task Performance}

\subsection{Parametric Functions}
\label{sec:appendix_parametric_scaling_functions}

\begin{table*}[h]
    \centering \small
    \begin{tabular}{@{}l@{\hspace{1em}}l@{\hspace{1em}}l@{\hspace{1em}}c@{}}
        \toprule
        \multicolumn{1}{l}{\textbf{Form}} & \multicolumn{2}{c}{\textbf{Expressions for $f(x, \finetunedatasize)$}} & \textbf{$x^*(\computebudget)$ ind. $\finetunedatasize$} \\
        \cmidrule(l){2-3}
        & Multiple Factors ($K>1$) & Single Factor ($K=1$) & \\
        \midrule
        \texttt{mult} & $\alpha (\prod_k x_k^{-a_k}) \finetunedatasize^{-\datascalingexponent} + \varepsilon$ & $\alpha x^{-a} \finetunedatasize^{-\datascalingexponent} + \varepsilon$ & \greencheck \\
        \texttt{add} & $\sum_k \alpha_k x_k^{-a_k} + \xi \finetunedatasize^{-\datascalingexponent} + \varepsilon$ & $\alpha x^{-a} + \xi \finetunedatasize^{-\datascalingexponent} + \varepsilon$ & \greencheck \\
        \texttt{add-interact$_{s}$} & $\sum_k \alpha_k x_k^{-a_k} + \sum_k \beta_k x_k^{b_k} \finetunedatasize^{-\datascalingexponent} + \varepsilon$ & $\alpha x^{-a} + \beta x^b \finetunedatasize^{-\datascalingexponent} + \varepsilon$ & \redcross \\
        \texttt{add-interact} & $\sum_k \alpha_k x_k^{-a_k} + \sum_k \beta_k x_k^{b_k} \finetunedatasize^{-\datascalingexponent} + \xi \finetunedatasize^{-\datascalingexponent} + \varepsilon$ & $\alpha x^{-a} + \beta x^b \finetunedatasize^{-\datascalingexponent} + \xi \finetunedatasize^{-\datascalingexponent} + \varepsilon$ & \redcross \\
        \bottomrule
    \end{tabular}
    \caption{
        \textbf{Parametric Models of Task Performance.} This table summarizes the parametric functional forms used to model task performance, distinguishing between cases where all scaling factors $x$ are modeled jointly with the finetuning data size $\finetunedatasize$ ($K>1$) and cases where each scaling factor is modeled individually with $\finetunedatasize$ ($K=1$). The \texttt{add-interact} form combines additive power laws with interaction terms between $x_k$ and $\finetunedatasize$. \texttt{add-interact$_{s}$} simplifies this by removing the standalone term dependent on $\finetunedatasize$. 
        \texttt{add} represents a standard additive power law, while \texttt{mult} corresponds to a multiplicative power law. Coefficients (e.g., $\alpha, \beta, \xi, \varepsilon$) and scaling exponents (e.g., $a, b, \datascalingexponent$) are denoted by Greek and English letters, respectively. The final column indicates whether varying $\finetunedatasize$ affects the optimal scaling factors that is the solution to the optimization problem defined in Equation~\ref{eq:inference_compute_optimal_scaling_opt_problem}.
    }
    \label{tab:scaling_functional_form_expressions}
\end{table*}

We investigate several parametric functions to model task error in video VLMs as a function of scaling factors $x$ and finetuning data size $\finetunedatasize$. These functions capture the diminishing returns observed in scaling laws, where increasing $x$ or $\finetunedatasize$ yields progressively smaller performance gains. For example, adding more frames or increasing the finetuning data size improves performance initially, but the benefit diminishes as the model saturates in available information or data. This behavior parallels the concept of diminishing marginal utility in economics.

The primary function we use is the \texttt{add-interact} form, which combines additive power-law terms with interaction terms between each scaling factor $x_k$ and $\finetunedatasize$. This formulation is particularly effective for analyzing the joint influence of scaling factors and data size on performance under constraints on inference compute $c(x)$ and data size $\finetunedatasize$. This approach was first proposed to study the relationship between architecture factors (e.g., width, depth) and pretraining data size when scaling ViT models \citep{alabdulmohsinGettingViTShape2023}. A simplified variant, \texttt{add-interact$_{s}$}, removes the standalone term dependent on $\finetunedatasize$. We also evaluate the \texttt{add} model, a special case of \texttt{add-interact} that assumes independent contributions from $x$ and $\finetunedatasize$. This additive power-law model is widely used in pretraining scaling law studies \citep{kaplanScalingLawsNeural2020, hoffmannTrainingComputeoptimalLarge2022} and has been adapted for finetuning scenarios \citep{duExploringDesignSpace2024}. Additionally, we consider the \texttt{mult} model, a multiplicative power-law formulation that is sometimes preferred in finetuning scaling law studies \citep{weiFinetunedLanguageModels2022, liInferenceOptimalVLMs2024}.

Table~\ref{tab:scaling_functional_form_expressions} provides the expressions for each parametric function. By comparing these formulations, we aim to identify the most effective model for capturing the scaling behavior of video VLMs.

\newpage
\subsection{What is the effect of exponent parameter \texorpdfstring{$b$}{b} in \texttt{add-interact} parametric function?}
\label{sec:interpret_exponent_b_in_add_interact_parametric_form}

We analyze the role of the exponent parameter $b \in \mathbb{R}$ in the \texttt{add-interact} parametric form. For simplicity, we redefine the function for a scalar $x$ as
\begin{align}
    f(x, \finetunedatasize) = \alpha x^{-a} + (\beta x^b + \xi) \finetunedatasize^{-\datascalingexponent} + \varepsilon.
\end{align}
The parameter $b$ controls how the scaling factor $x$ influences the coefficient $(\beta x^b + \xi)$, which represents the reducible error associated with scaling the finetuning data size $\finetunedatasize$. A positive $b > 0$ has two key effects:

\begin{enumerate}
    \item \textbf{Larger $x$ Requires More Data to Match the Same Error:} When $b > 0$, the coefficient $(\beta x^b + \xi)$ increases with $x$. For two values $x_{\text{S}} < x_{\text{L}}$, we have $(\beta x_{\text{S}}^b + \xi) < (\beta x_{\text{L}}^b + \xi)$. To achieve the same error $f(x, \finetunedatasize)$, a larger $x_{\text{L}}$ requires a larger finetuning data size $\finetunedatasize_{\text{L}}$ compared to $\finetunedatasize_{\text{S}}$. Specifically, the equality $(\beta x_{\text{S}}^b + \xi) \finetunedatasize_{\text{S}}^{-\datascalingexponent} = (\beta x_{\text{L}}^b + \xi) \finetunedatasize_{\text{L}}^{-\datascalingexponent}$ holds only if $\finetunedatasize_{\text{S}} < \finetunedatasize_{\text{L}}$. Thus, $b > 0$ increases the data required for larger $x$ to match the performance of smaller $x$.

    \item \textbf{Larger $x$ Reduces Error Faster Per Example:} A positive $b$ also increases the marginal benefit of finetuning data for larger $x$. The derivative of the reducible error term with respect to $\finetunedatasize$ is $\frac{\partial}{\partial \finetunedatasize} (\beta x^b + \xi) \finetunedatasize^{-\datascalingexponent} = -\datascalingexponent (\beta x^b + \xi) \finetunedatasize^{-\datascalingexponent-1}$. For larger $x$, the term $(\beta x^b + \xi)$ is larger, making the derivative more negative. This implies that the absolute rate of error reduction (i.e., the decrease in error per additional finetuning example) is greater for larger $x$ when $b > 0$. As a result, $b > 0$ improves the efficiency of finetuning for larger $x$.
\end{enumerate}

In contrast, when $b < 0$, the effects are reversed: larger $x$ requires less data to achieve the same error, but the marginal benefit of each additional finetuning example decreases.

\newpage
\subsection{Fitting the Parametric Model}
\label{sec:parametric_model_fitting_ablations}

\begin{table}[h]
    \centering \scriptsize
    \begin{minipage}[t]{0.49\textwidth}
    \centering
    \begin{tabular}{llp{.65cm}p{.65cm}p{.7cm}p{.65cm}p{.65cm}}
        \toprule
        \textbf{loss} & \multicolumn{3}{c}{\textbf{Star+IsoFLOP CV{\scriptsize (5-fold)}}} & \multicolumn{3}{c}{\textbf{Star $\rightarrow$ IsoFLOP}} \\
         & MSE {\scriptsize $\downarrow$} & $E_{\%}$ {\scriptsize $\downarrow$} & $R^2$ {\scriptsize $\uparrow$} & MSE {\scriptsize $\downarrow$} & $E_{\%}$ {\scriptsize $\downarrow$} & $R^2$ {\scriptsize $\uparrow$} \\
         \midrule
         Huber  & 0.44 & 0.97 & 0.95 & 1.11 & 1.38 & 0.91 \\
         MSE & \cellcolor[rgb]{0.30196078431372547,0.7215686274509804,1.0}0.2 & \cellcolor[rgb]{0.30196078431372547,0.7215686274509804,1.0}0.77 & \cellcolor[rgb]{0.30196078431372547,0.7215686274509804,1.0}0.98 & \cellcolor[rgb]{0.30196078431372547,0.7215686274509804,1.0}0.95 & \cellcolor[rgb]{0.30196078431372547,0.7215686274509804,1.0}1.33 & \cellcolor[rgb]{0.30196078431372547,0.7215686274509804,1.0}0.92 \\
        \bottomrule
        \end{tabular}
    \end{minipage}
    \begin{minipage}[t]{0.49\textwidth}
        \centering
        \begin{tabular}{llp{.65cm}p{.65cm}p{.7cm}p{.65cm}p{.65cm}}
        \toprule
        \textbf{init} & \multicolumn{3}{c}{\textbf{Star+IsoFLOP CV{\scriptsize (5-fold)}}} & \multicolumn{3}{c}{\textbf{Star $\rightarrow$ IsoFLOP}} \\
            & MSE {\scriptsize $\downarrow$} & $E_{\%}$ {\scriptsize $\downarrow$} & $R^2$ {\scriptsize $\uparrow$} & MSE {\scriptsize $\downarrow$} & $E_{\%}$ {\scriptsize $\downarrow$} & $R^2$ {\scriptsize $\uparrow$} \\
        \midrule
        \texttt{zero} & 0.43 & 0.99 & 0.95 & 1.36 & 1.63 & 0.89 \\
        \texttt{r(100)} & 0.34 & 0.88 & 0.96 & 0.96 & 1.36 & \cellcolor[rgb]{0.30196078431372547,0.7215686274509804,1.0}0.92 \\
        \texttt{r(500)} & \cellcolor[rgb]{0.30196078431372547,0.7215686274509804,1.0}0.2 & \cellcolor[rgb]{0.30196078431372547,0.7215686274509804,1.0}0.77 & \cellcolor[rgb]{0.30196078431372547,0.7215686274509804,1.0}0.98 & \cellcolor[rgb]{0.30196078431372547,0.7215686274509804,1.0}0.95 & \cellcolor[rgb]{0.30196078431372547,0.7215686274509804,1.0}1.33 & \cellcolor[rgb]{0.30196078431372547,0.7215686274509804,1.0}0.92 \\
        \bottomrule
        \end{tabular}
    \end{minipage}
    \begin{minipage}[t]{0.49\textwidth}
        \centering
        \begin{tabular}{llp{.65cm}p{.65cm}p{.7cm}p{.65cm}p{.65cm}}
        \toprule
        \textbf{bound} & \multicolumn{3}{c}{\textbf{Star+IsoFLOP CV{\scriptsize (5-fold)}}} & \multicolumn{3}{c}{\textbf{Star $\rightarrow$ IsoFLOP}} \\
            & MSE {\scriptsize $\downarrow$} & $E_{\%}$ {\scriptsize $\downarrow$} & $R^2$ {\scriptsize $\uparrow$} & MSE {\scriptsize $\downarrow$} & $E_{\%}$ {\scriptsize $\downarrow$} & $R^2$ {\scriptsize $\uparrow$} \\
        \midrule
        \greencheck & 0.57 & 1.11 & 0.94 & 2.45 & 2.42 & 0.8 \\
        \redcross & \cellcolor[rgb]{0.30196078431372547,0.7215686274509804,1.0}0.2 & \cellcolor[rgb]{0.30196078431372547,0.7215686274509804,1.0}0.77 & \cellcolor[rgb]{0.30196078431372547,0.7215686274509804,1.0}0.98 & \cellcolor[rgb]{0.30196078431372547,0.7215686274509804,1.0}0.95 & \cellcolor[rgb]{0.30196078431372547,0.7215686274509804,1.0}1.33 & \cellcolor[rgb]{0.30196078431372547,0.7215686274509804,1.0}0.92 \\
        \bottomrule
        \end{tabular}
    \end{minipage}
    \begin{minipage}[t]{0.49\textwidth}
        \centering
        \begin{tabular}{llp{.65cm}p{.65cm}p{.7cm}p{.65cm}p{.65cm}}
        \toprule
        \textbf{log space} & \multicolumn{3}{c}{\textbf{Star+IsoFLOP CV{\scriptsize (5-fold)}}} & \multicolumn{3}{c}{\textbf{Star $\rightarrow$ IsoFLOP}} \\
         & MSE {\scriptsize $\downarrow$} & $E_{\%}$ {\scriptsize $\downarrow$} & $R^2$ {\scriptsize $\uparrow$} & MSE {\scriptsize $\downarrow$} & $E_{\%}$ {\scriptsize $\downarrow$} & $R^2$ {\scriptsize $\uparrow$} \\
         \midrule
         \redcross & 0.29 & 0.88 & 0.97 & 19.72 & 2.93 & -0.62 \\
         \greencheck & \cellcolor[rgb]{0.30196078431372547,0.7215686274509804,1.0}0.2 & \cellcolor[rgb]{0.30196078431372547,0.7215686274509804,1.0}0.77 & \cellcolor[rgb]{0.30196078431372547,0.7215686274509804,1.0}0.98 & \cellcolor[rgb]{0.30196078431372547,0.7215686274509804,1.0}0.95 & \cellcolor[rgb]{0.30196078431372547,0.7215686274509804,1.0}1.33 & \cellcolor[rgb]{0.30196078431372547,0.7215686274509804,1.0}0.92 \\
        \bottomrule
        \end{tabular}
    \end{minipage}
    \caption{
        \textbf{Ablation Study on Fitting Procedures.} We analyze the impact of key design choices for fitting \texttt{add-interact} to predict average task performance across two setups: (1) \textit{in-distribution} evaluation using 5-fold cross-validation (CV) on combined star and isoFLOP data, and (2) \textit{extrapolation} on isoFLOP data after training on star data. Metrics include mean squared error (MSE), average relative error ($E_{\%}$), and coefficient of determination ($R^2$). The ablations include: (1) \textit{loss function} (top left): MSE provides better fit than Huber loss, (2) \textit{parameter initialization} (top right): e.g., \texttt{r(500)} initializes parameters randomly within a range and selects the best model from 500 runs, while \texttt{zero} sets all parameters to zero, (3) \textit{positivity constraint} (bottom left): enforcing positivity for the exponents degrades performance, and (4) \textit{log-space computation} (bottom right): transforming the scaling function into log space improves numerical stability. Results highlight the importance of careful parameter initialization and log-space computation, while positivity constraints are detrimental, and both loss functions perform comparably.
    }
    \label{tab:ablate_scaling_fitting_setups}
\end{table}

To estimate the parameters $\theta$ of the parametric model of task performance $f(x, \finetunedatasize)$, we minimize the relative error between predicted and observed performance in log space:
\begin{align}
    \min_{\theta} \sum_{\text{Run } i} \left( \log f(x^{(i)}, n^{(i)}; \theta) - \log f^{(i)} \right)^2.
\end{align}
We solve the above optimization problem using \texttt{scipy}'s implementation of the L-BFGS algorithm. Table~\ref{tab:ablate_scaling_fitting_setups} presents an ablation study on key design choices for fitting the parametric model. Below, we summarize the findings:
\begin{itemize}
    \item \textbf{Loss Function}: We compare mean squared error (MSE) and Huber loss. MSE consistently outperforms Huber loss and is therefore used as the objective.
    \item \textbf{Parameter Initialization}: The optimization problem is nonconvex and nonlinear, so we mitigate the risk of poor local minima by running 500 trials with random initializations and selecting the best fit. Coefficient parameters (e.g., $\alpha, \beta, \xi, \varepsilon$) are sampled uniformly from $(0, 30)$, while exponent parameters (e.g., $a, b, d$) are sampled from $(-1, 1)$.
    \item \textbf{Positivity Constraints}: Unlike \citet{alabdulmohsinGettingViTShape2023}, we do not enforce positivity constraints on exponent parameters. Ablation results show that such constraints degrade performance.
    \item \textbf{Log-Space Computation}: Following \citet{hoffmannTrainingComputeoptimalLarge2022}, we compute $f(x, n)$ in log space. This improves numerical stability and yields better performance across all metrics.
\end{itemize}
In summary, we fit the parametric models by minimizing relative error in log space using MSE loss. Parameters are initialized randomly, and the best fit is selected from 500 runs. We avoid positivity constraints on exponents and leverage log-space computation for improved stability and accuracy.

\newpage
\subsection{Bootstrap Aggregation}
\label{sec:appendix_bootstrap_aggregation}
    
\begin{table*}[h]
    \centering \small
    \begin{tabular}{llp{.85cm}p{.85cm}p{.85cm}p{.85cm}p{.85cm}p{.85cm}}
        \toprule
        \textbf{\# Resamples} & \textbf{Aggregation} & \multicolumn{3}{c}{\textbf{Train}} & \multicolumn{3}{c}{\textbf{Star $\rightarrow$ IsoFLOP}} \\
        &  & MSE {\scriptsize $\downarrow$} & $E_{\%}$ {\scriptsize $\downarrow$} & $R^2$ {\scriptsize $\uparrow$} & MSE {\scriptsize $\downarrow$} & $E_{\%}$ {\scriptsize $\downarrow$} & $R^2$ {\scriptsize $\uparrow$} \\
        \midrule
        100 & mean & \cellcolor[rgb]{0.6,0.8392156862745098,1.0}0.15 & \cellcolor[rgb]{0.9019607843137255,0.9529411764705882,1.0}0.58 & \cellcolor[rgb]{0.30196078431372547,0.7215686274509804,1.0}0.99 & 4418 & 28 & \cellcolor[rgb]{0.9019607843137255,0.9529411764705882,1.0}-362 \\
        1 & median & \cellcolor[rgb]{0.9019607843137255,0.9529411764705882,1.0}0.17 & 0.62 & \cellcolor[rgb]{0.6,0.8392156862745098,1.0}0.98 & \cellcolor[rgb]{0.9019607843137255,0.9529411764705882,1.0}1.24 & \cellcolor[rgb]{0.9019607843137255,0.9529411764705882,1.0}1.55 & \cellcolor[rgb]{0.6,0.8392156862745098,1.0}0.9 \\
        50 & median & \cellcolor[rgb]{0.30196078431372547,0.7215686274509804,1.0}0.14 & \cellcolor[rgb]{0.6,0.8392156862745098,1.0}0.57 & \cellcolor[rgb]{0.30196078431372547,0.7215686274509804,1.0}0.99 & \cellcolor[rgb]{0.6,0.8392156862745098,1.0}0.96 & \cellcolor[rgb]{0.6,0.8392156862745098,1.0}1.35 & \cellcolor[rgb]{0.30196078431372547,0.7215686274509804,1.0}0.92 \\
        100 & median & \cellcolor[rgb]{0.30196078431372547,0.7215686274509804,1.0}0.14 & \cellcolor[rgb]{0.30196078431372547,0.7215686274509804,1.0}0.56 & \cellcolor[rgb]{0.30196078431372547,0.7215686274509804,1.0}0.99 & \cellcolor[rgb]{0.30196078431372547,0.7215686274509804,1.0}0.95 & \cellcolor[rgb]{0.30196078431372547,0.7215686274509804,1.0}1.33 & \cellcolor[rgb]{0.30196078431372547,0.7215686274509804,1.0}0.92 \\
        \bottomrule
    \end{tabular}
    \caption{
        \textbf{Impact of Bootstrap Aggregation Design Choices.} This table evaluates the effect of bootstrap aggregation strategies when using \texttt{add-interact} to predict average task performance on training results from the isoFLOP sweep when trained on that is the star sweep. Metrics include mean squared error (MSE), average relative error ($E_{\%}$), and coefficient of determination ($R^2$). Results show that median aggregation consistently outperforms mean aggregation. Performance improves with more bootstrap resamples (or fitted base models), with the best results achieved using median aggregation with 100 resamples that we adopt.
    }
    \label{tab:ablate_bagged_model_setup}
\end{table*}

Figure~\ref{fig:parametric_fitting_of_task_performance_several_aspect_summary} (left) highlights the high variance in parameter estimates for the parametric model of task performance when trained on approximately 100 examples. This variance poses a significant challenge due to the limited data. To mitigate this, we apply bootstrap aggregation (bagging), where multiple base models are trained on bootstrap-resampled datasets, and their predictions are aggregated.

Table~\ref{tab:ablate_bagged_model_setup} summarizes the ablation study on the bagging setup, analyzing the effects of the number of resamples and the aggregation method on model performance. The results demonstrate that median aggregation consistently outperforms mean aggregation across all metrics. Furthermore, increasing the number of resamples improves performance, with the best results obtained using 100 resamples and median aggregation. This setup significantly improves goodness-of-fit, and it is adopted in our work.

\newpage
\section[Measuring Sensitivity of Compute-Optimal Frontier to Data Size]{Measuring Sensitivity of Compute-Optimal Frontier $x^*(\computebudget, \finetunedatasize)$ to Data Size $\finetunedatasize$}
\label{sec:appendix_details_on_elasticity_and_ablation_on_computation}

\begin{table}[h]
    \centering \scriptsize
    \begin{minipage}[t]{0.24\textwidth}
    \centering
    \begin{tabular}{llll}
        \toprule
        $|\mathsf{C}|$ & $e_N$ & $e_T$ & $e_V$ \\
        \midrule
        1 & 0.0 & \cellcolor[rgb]{0.9019607843137255,0.9529411764705882,1.0}0.0 & 0.0 \\
        10 & \cellcolor[rgb]{0.9019607843137255,0.9529411764705882,1.0}-0.19 & \cellcolor[rgb]{0.30196078431372547,0.7215686274509804,1.0}0.17 & 0.58 \\
        50 & \cellcolor[rgb]{0.30196078431372547,0.7215686274509804,1.0}-0.23 & \cellcolor[rgb]{0.6,0.8392156862745098,1.0}0.16 & \cellcolor[rgb]{0.30196078431372547,0.7215686274509804,1.0}0.89 \\
        100 & \cellcolor[rgb]{0.30196078431372547,0.7215686274509804,1.0}-0.23 & \cellcolor[rgb]{0.6,0.8392156862745098,1.0}0.16 & \cellcolor[rgb]{0.6,0.8392156862745098,1.0}0.88 \\
        200 & \cellcolor[rgb]{0.6,0.8392156862745098,1.0}-0.22 & \cellcolor[rgb]{0.30196078431372547,0.7215686274509804,1.0}0.17 & 0.78 \\
        300 & \cellcolor[rgb]{0.6,0.8392156862745098,1.0}-0.22 & \cellcolor[rgb]{0.30196078431372547,0.7215686274509804,1.0}0.17 & \cellcolor[rgb]{0.9019607843137255,0.9529411764705882,1.0}0.79 \\
        \bottomrule
        \end{tabular}
    \end{minipage}
    \begin{minipage}[t]{0.24\textwidth}
    \centering
    \begin{tabular}{llll}
        \toprule
        $|\mathsf{N}|$ & $e_N$ & $e_T$ & $e_V$ \\
        \midrule
        1 & \cellcolor[rgb]{0.6,0.8392156862745098,1.0}-0.04 & \cellcolor[rgb]{0.6,0.8392156862745098,1.0}0.05 & 0.12 \\
        10 & \cellcolor[rgb]{0.30196078431372547,0.7215686274509804,1.0}-0.22 & \cellcolor[rgb]{0.30196078431372547,0.7215686274509804,1.0}0.17 & \cellcolor[rgb]{0.9019607843137255,0.9529411764705882,1.0}0.77 \\
        20 & \cellcolor[rgb]{0.30196078431372547,0.7215686274509804,1.0}-0.22 & \cellcolor[rgb]{0.30196078431372547,0.7215686274509804,1.0}0.17 & \cellcolor[rgb]{0.6,0.8392156862745098,1.0}0.78 \\
        50 & \cellcolor[rgb]{0.30196078431372547,0.7215686274509804,1.0}-0.22 & \cellcolor[rgb]{0.30196078431372547,0.7215686274509804,1.0}0.17 & \cellcolor[rgb]{0.30196078431372547,0.7215686274509804,1.0}0.79 \\
        100 & \cellcolor[rgb]{0.30196078431372547,0.7215686274509804,1.0}-0.22 & \cellcolor[rgb]{0.30196078431372547,0.7215686274509804,1.0}0.17 & \cellcolor[rgb]{0.30196078431372547,0.7215686274509804,1.0}0.79 \\
        \bottomrule
        \end{tabular}
    \end{minipage}
    \begin{minipage}[t]{0.24\textwidth}
    \centering
    \begin{tabular}{llll}
        \toprule
        $\max(\mathsf{N})$ & $e_N$ & $e_T$ & $e_V$ \\
        \midrule
        1 & -0.04 & 0.05 & 0.12 \\
        2 & -0.06 & 0.08 & 0.19 \\
        4 & -0.12 & 0.12 & 0.42 \\
        6 & \cellcolor[rgb]{0.9019607843137255,0.9529411764705882,1.0}-0.16 & \cellcolor[rgb]{0.9019607843137255,0.9529411764705882,1.0}0.13 & \cellcolor[rgb]{0.9019607843137255,0.9529411764705882,1.0}0.58 \\
        8 & \cellcolor[rgb]{0.6,0.8392156862745098,1.0}-0.19 & \cellcolor[rgb]{0.6,0.8392156862745098,1.0}0.15 & \cellcolor[rgb]{0.6,0.8392156862745098,1.0}0.7 \\
        10 & \cellcolor[rgb]{0.30196078431372547,0.7215686274509804,1.0}-0.22 & \cellcolor[rgb]{0.30196078431372547,0.7215686274509804,1.0}0.17 & \cellcolor[rgb]{0.30196078431372547,0.7215686274509804,1.0}0.79 \\
        \bottomrule
        \end{tabular}
    \end{minipage}
    \begin{minipage}[t]{0.24\textwidth}
    \centering
    \begin{tabular}{llll}
        \toprule
        $\triangle n$ & $e_N$ & $e_T$ & $e_V$ \\
        \midrule
        0.1 & 0.32 & \cellcolor[rgb]{0.30196078431372547,0.7215686274509804,1.0}0.2 & \cellcolor[rgb]{0.30196078431372547,0.7215686274509804,1.0}1.13 \\
        0.5 & -0.08 & \cellcolor[rgb]{0.30196078431372547,0.7215686274509804,1.0}0.2 & \cellcolor[rgb]{0.6,0.8392156862745098,1.0}0.99 \\
        1.0 & -0.15 & \cellcolor[rgb]{0.30196078431372547,0.7215686274509804,1.0}0.2 & \cellcolor[rgb]{0.9019607843137255,0.9529411764705882,1.0}0.96 \\
        2.0 & -0.2 & \cellcolor[rgb]{0.6,0.8392156862745098,1.0}0.19 & 0.89 \\
        3.0 & \cellcolor[rgb]{0.30196078431372547,0.7215686274509804,1.0}-0.23 & \cellcolor[rgb]{0.9019607843137255,0.9529411764705882,1.0}0.18 & 0.83 \\
        4.0 & \cellcolor[rgb]{0.30196078431372547,0.7215686274509804,1.0}-0.23 & 0.17 & 0.8 \\
        5.0 & \cellcolor[rgb]{0.6,0.8392156862745098,1.0}-0.22 & 0.17 & 0.79 \\
        6.0 & \cellcolor[rgb]{0.6,0.8392156862745098,1.0}-0.22 & 0.17 & 0.78 \\
        7.0 & \cellcolor[rgb]{0.6,0.8392156862745098,1.0}-0.22 & 0.16 & 0.75 \\
        8.0 & \cellcolor[rgb]{0.9019607843137255,0.9529411764705882,1.0}-0.21 & 0.16 & 0.72 \\
        9.0 & -0.2 & 0.15 & 0.68 \\
        10.0 & -0.19 & 0.15 & 0.64 \\
        \bottomrule
        \end{tabular}
    \end{minipage}
    \caption{
        \textbf{Ablation Study on Elasticity Computation.}This table evaluates the impact of key parameters on the computation of elasticity, including: (1) the number of inference compute budgets $|\mathsf{C}|$ (measured in TFLOPs) used for averaging, (2) the number of finetuning data sizes $|\mathsf{N}|$ (measured in millions) used for averaging, (3) the maximum finetuning data size $\max(\mathsf{N})$ (upper bound of data sizes in $\mathsf{N}$), and (4) the step size $\triangle n$ (in millions) used for the forward difference approximation of the derivative. Results show that elasticity values stabilize with larger $|\mathsf{C}|$ and $|\mathsf{N}|$, so we set $|\mathsf{C}|=300$ and $|\mathsf{N}|=100$ to minimize variability. While $\max(\mathsf{N})$ affects the magnitude of elasticity, it preserves consistent trends across $k \in \{N, T, V\}$. We set $\max(\mathsf{N})=10$M to ensure broad coverage of data sizes. The step size $\triangle n$ affects the stability and precision of the derivative approximation; we select $\triangle n=5$ to balance this trade-off. These choices ensure reliable and interpretable elasticity estimates. 
    }
    \label{tab:ablation_study_to_compute_elasticity}
\end{table}

Figure~\ref{fig:plt_compute_optimal_scaling_of_factors_for_fitted_scaling_function} shows that the inference compute-optimal frontier $x_k^*(\computebudget)$, shifts up or down depending on the finetuning data size $\finetunedatasize$.  To quantify both the direction and magnitude of this shift, we define ``elasticity''
\begin{align}
    e_k(\computebudget, \finetunedatasize)
        = \frac{\partial  x_k^*(\computebudget; \finetunedatasize)}{\partial \finetunedatasize} \cdot \frac{\finetunedatasize}{x_k^*(\computebudget; \finetunedatasize)}
\end{align}
to measure the relative sensitivity of $x_k^*$ to variations in $\finetunedatasize$ for a given inference compute budget $\computebudget$.

We use elasticity instead of simple partial derivatives, e.g., $\partial x_k^*(\computebudget; \finetunedatasize) / \partial \finetunedatasize$, because partial derivatives are highly sensitive to the scale of the variables involved. In our case, the scaling factors $x$ and the data size $\finetunedatasize$ differ significantly in magnitude: for example, LM size $x_N$ is on the order of billions, $x_T$ \& $x_V$ are in the tens or hundreds, and $\finetunedatasize$ is in the millions. Elasticity allows us to express relationships in percentage terms, making comparisons more intuitive. For instance, an elasticity value of $e_T = 0.1$ means that a $1\%$ increase in $\finetunedatasize$ results in a $0.1\%$ increase in $x_T^*$. This approach simplifies the interpretation of results and enables easier comparison of the sensitivity of $x^*$ to changes in $\finetunedatasize$.

Elasticity is commonly used in economics to measure how changes in one variable affect another, e.g., price elasticity of demand quantifies how price changes influence the quantity of goods demanded. This quantity is derived by solving a utility maximization problem under (monetary) budget constraints, often yielding closed-form solutions that allow direct computation of elasticity. In contrast, our approach solves the inference compute allocation problem in Equation~$ \ref{eq:inference_compute_optimal_scaling_opt_problem} $ using discrete search, which does not yield a closed-form expression for elasticity. Instead, we approximate elasticity using forward differences:
\begin{align}
    e_k(\computebudget,\finetunedatasize)
        \approx \frac{x_k^*(\computebudget; \finetunedatasize + \triangle \finetunedatasize)}{\triangle \finetunedatasize} \cdot \frac{\finetunedatasize}{x_k^*(\computebudget; \finetunedatasize)}
\end{align}
where $\triangle \finetunedatasize$ is the step size. To evaluate the overall impact of $\finetunedatasize$ on $x_k^*$, we calculate the average elasticity across compute budgets as 
\begin{align}
    \label{eq:elasticity_definition_as_function_of_data_size}
    e_k(\finetunedatasize) 
        = \frac{1}{|\mathsf{C}|} \sum_{\computebudget \in \mathsf{C}} e_k(\computebudget, \finetunedatasize),
\end{align}
where $\mathsf{C}$ represents the set of inference budgets of interest. We aggregate elasticity over both compute budgets and data sizes as 
\begin{align}
    \label{eq:elasticity_definition_as_function_of_data_size_and_compute_budget}
    e_k 
        = \frac{1}{|\mathsf{C}| |\mathsf{N}|} \sum_{\computebudget \in \mathsf{C}} \sum_{\finetunedatasize \in \mathsf{N}} e_k(\computebudget, \finetunedatasize),
\end{align}
with $\mathsf{N}$ denoting the set of finetuning data sizes of interest. These aggregated metrics provide a concise summary of how data size influences optimal scaling factors.

Table~\ref{tab:ablation_study_to_compute_elasticity} presents an ablation study analyzing key design choices for computing elasticity. The findings are summarized as follows:
\begin{itemize}
    \item \textbf{Size of $\mathsf{C}$:} We vary $|\mathsf{C}|$ (evenly spaced between $1$ TFLOP and $100$ TFLOPs) and observe that elasticity values stabilize as $|\mathsf{C}|$ increases. To minimize sensitivity to $|\mathsf{C}|$, we set $|\mathsf{C}|=300$.
    \item \textbf{Size of $\mathsf{N}$:} We vary $|\mathsf{N}|$ (evenly spaced between $1$M and $10$M) and find that elasticity values stabilize with larger $|\mathsf{N}|$. To reduce variability, we set $|\mathsf{N}|=100$.
    \item \textbf{Maximum Data Size $\max(\mathsf{N})$:} Ablating $\max(\mathsf{N})$ (with $\mathsf{N}$ sampled evenly starting from $1$M examples) reveals that elasticity magnitudes increase consistently with scaling. We set $\max(\mathsf{N})=10$M to ensure a broad range of data sizes is covered.
    \item \textbf{Step Size $\triangle n$:} The step size $\triangle n$ significantly affects elasticity. Small $\triangle n$ (e.g., adding one example) leads to near-zero elasticity except at points of sudden jumps, making numerical approximations unstable and sensitive to $\mathsf{C}$ and $\mathsf{N}$. Larger $\triangle n$ reduces accuracy of the forward difference approximation of the derivative. We select $\triangle n=5$ to balance stability and precision.
\end{itemize}
The resulting setup ensure reliable elasticity estimates.

\newpage
\section{Additional Results}
\subsection{Task-Specific Interpretation of Training Run Results}

\begin{figure*}[ht]
  \centering
  \myincludegraphics[width=.48\textwidth]{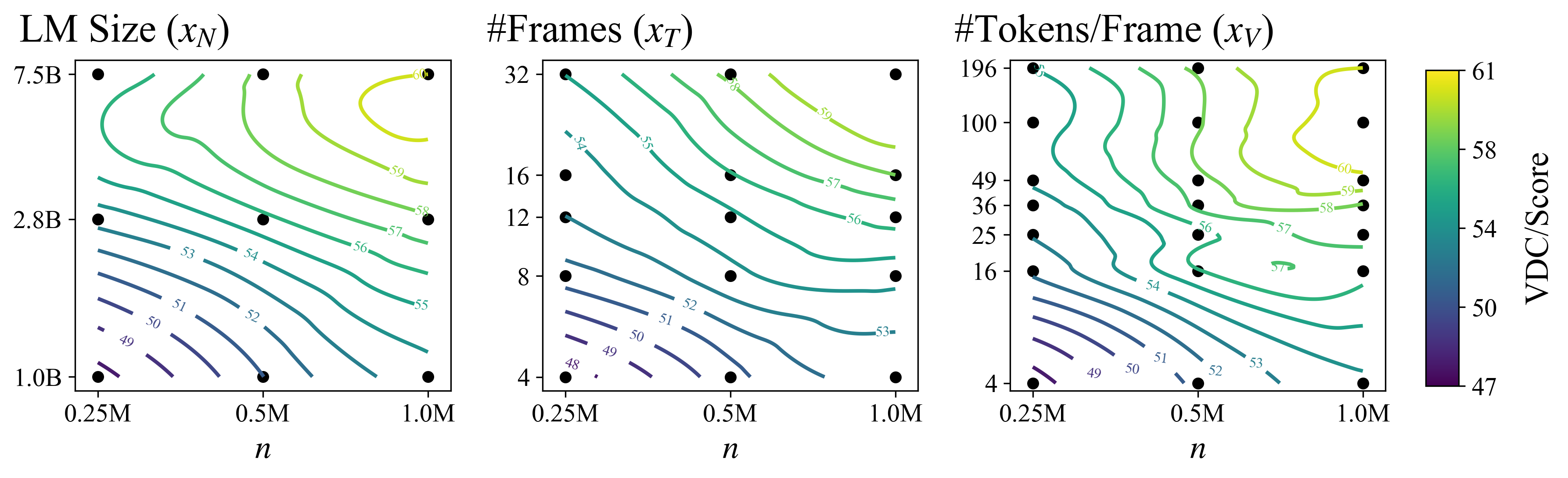}
  \myincludegraphics[width=.48\textwidth]{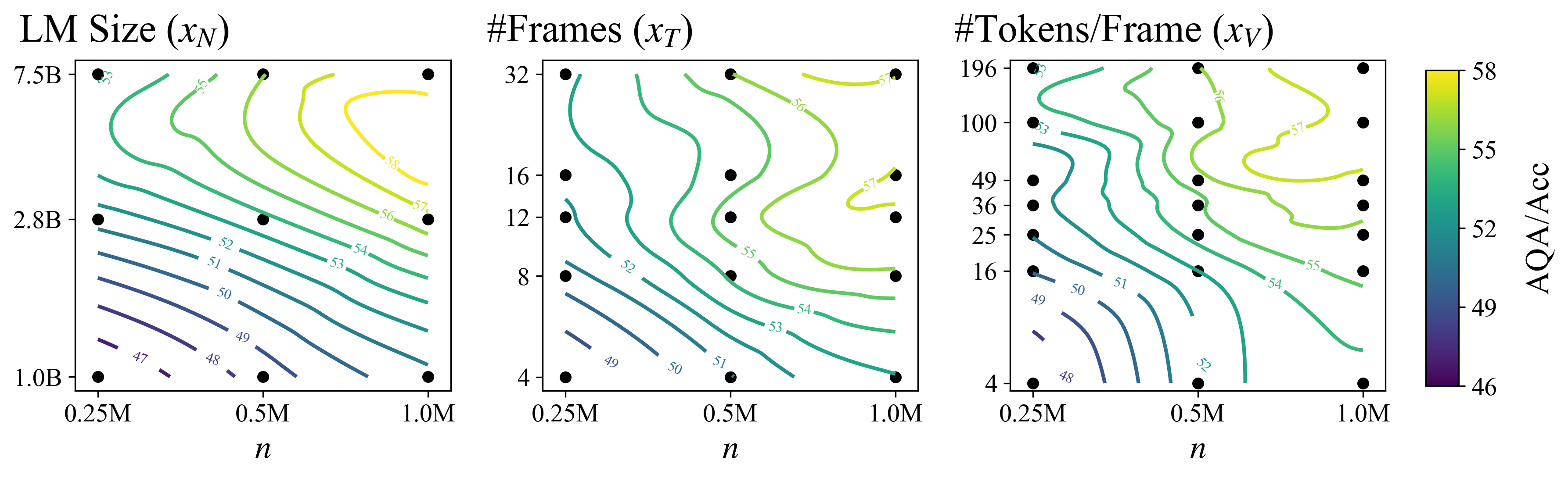}
  \myincludegraphics[width=.48\textwidth]{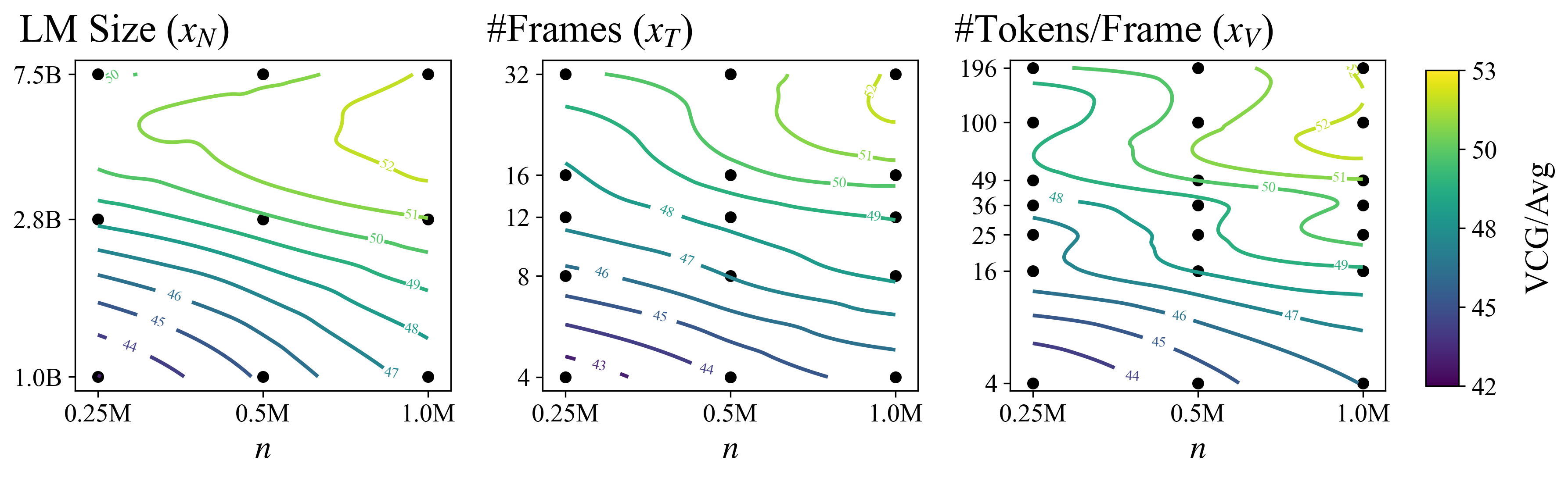}
  \myincludegraphics[width=.48\textwidth]{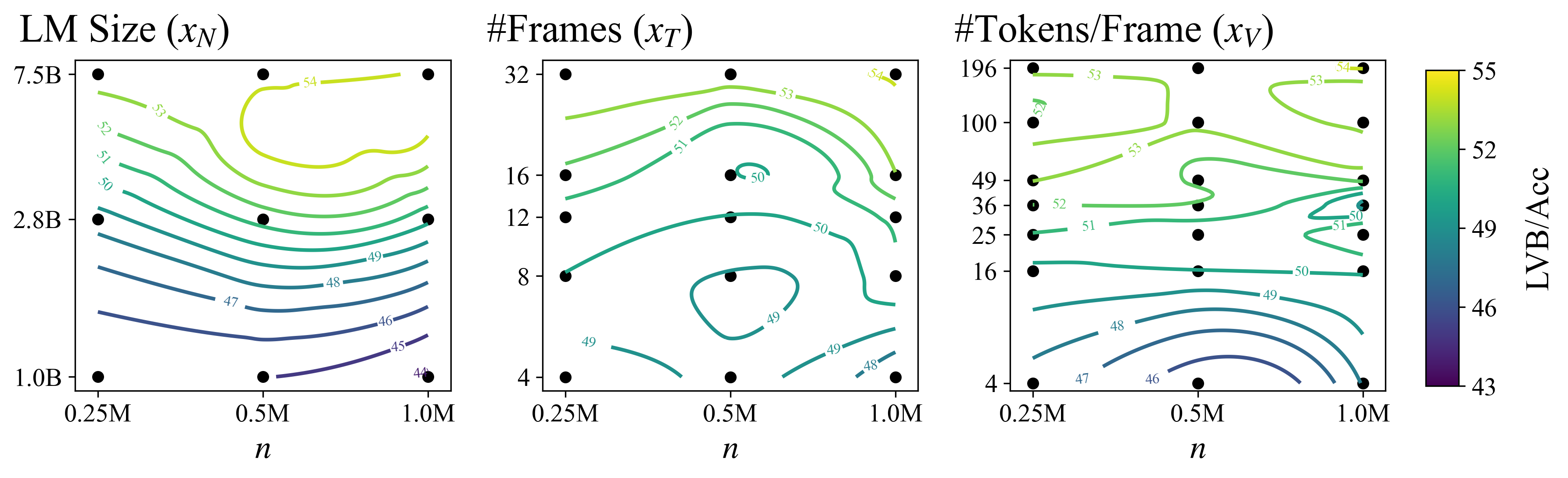}
  \myincludegraphics[width=.48\textwidth]{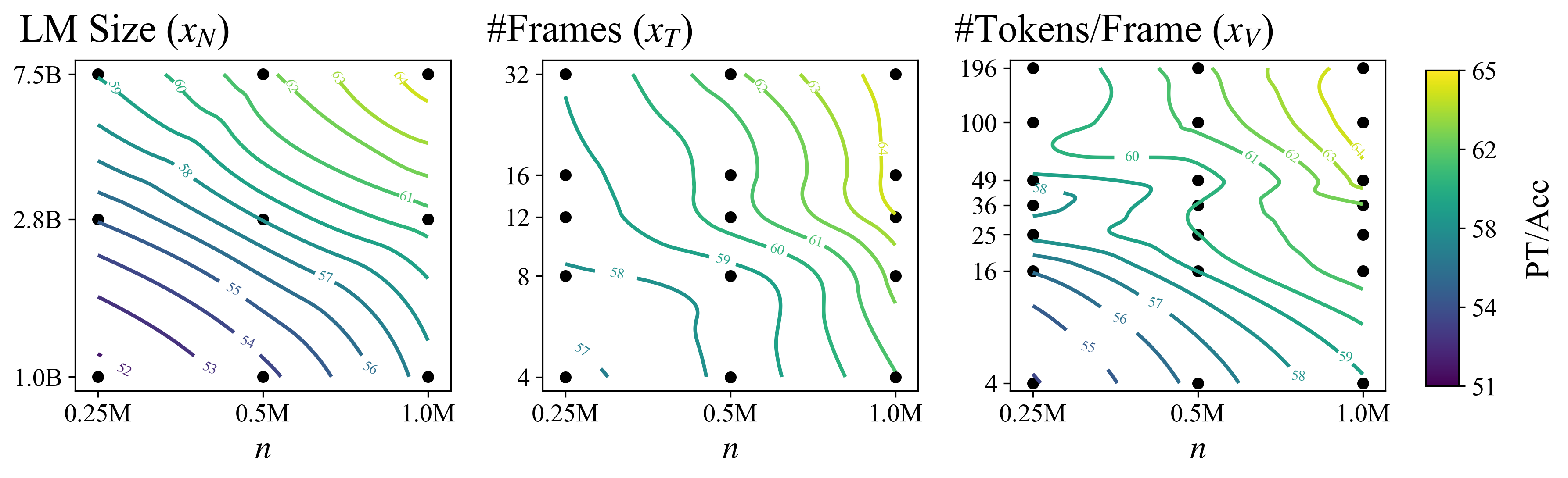}
  \myincludegraphics[width=.48\textwidth]{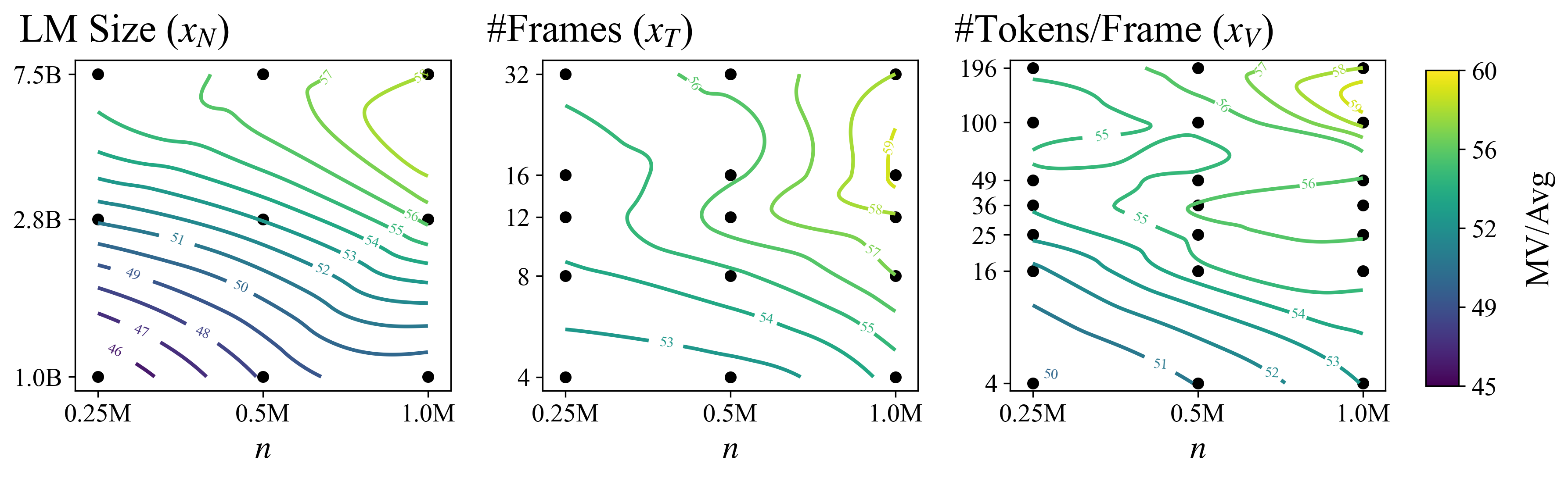}
  \myincludegraphics[width=.48\textwidth]{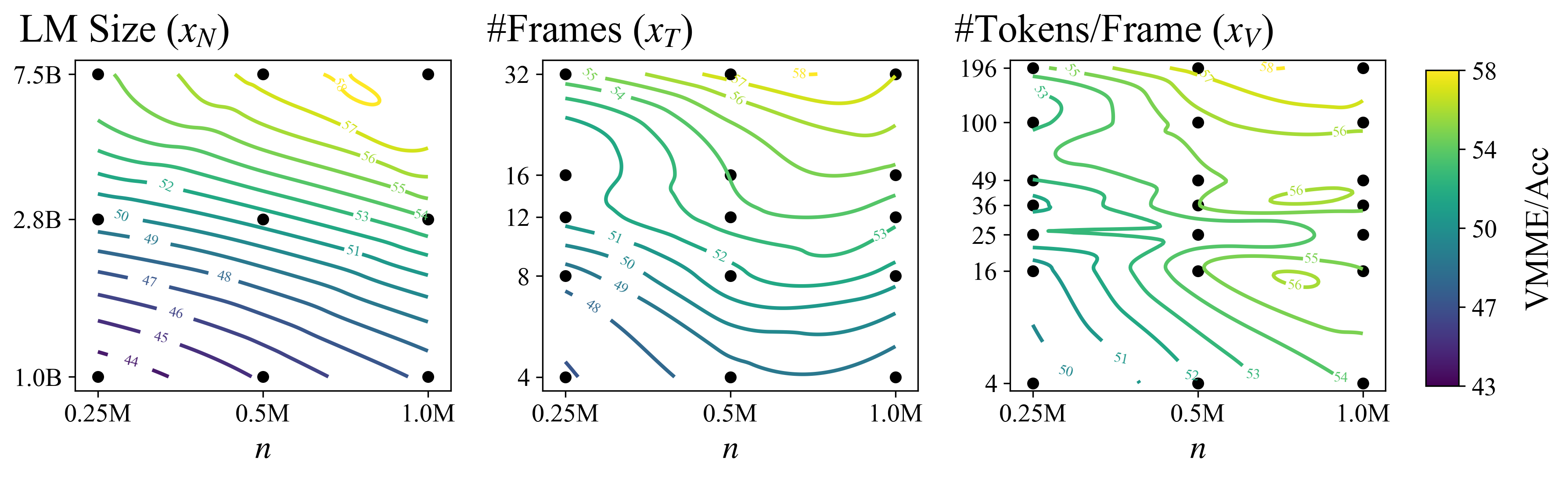}
  \myincludegraphics[width=.48\textwidth]{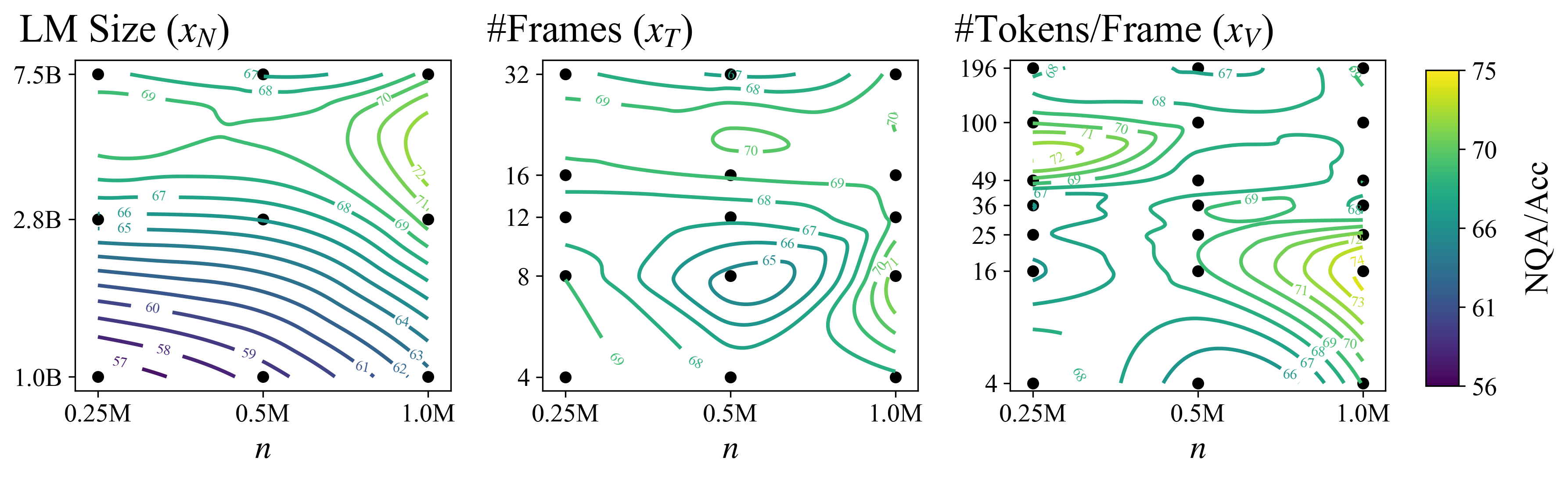}
  \caption{
    \textbf{Task-Specific IsoPerformance Contours.}
    }
  \label{fig:plt_star_sweep_performance_wrt_shape_param_and_data_size_logscale=both_metrics=all}
\end{figure*}

\begin{figure*}[ht]
    \centering
    \myincludegraphics[width=.48\textwidth]{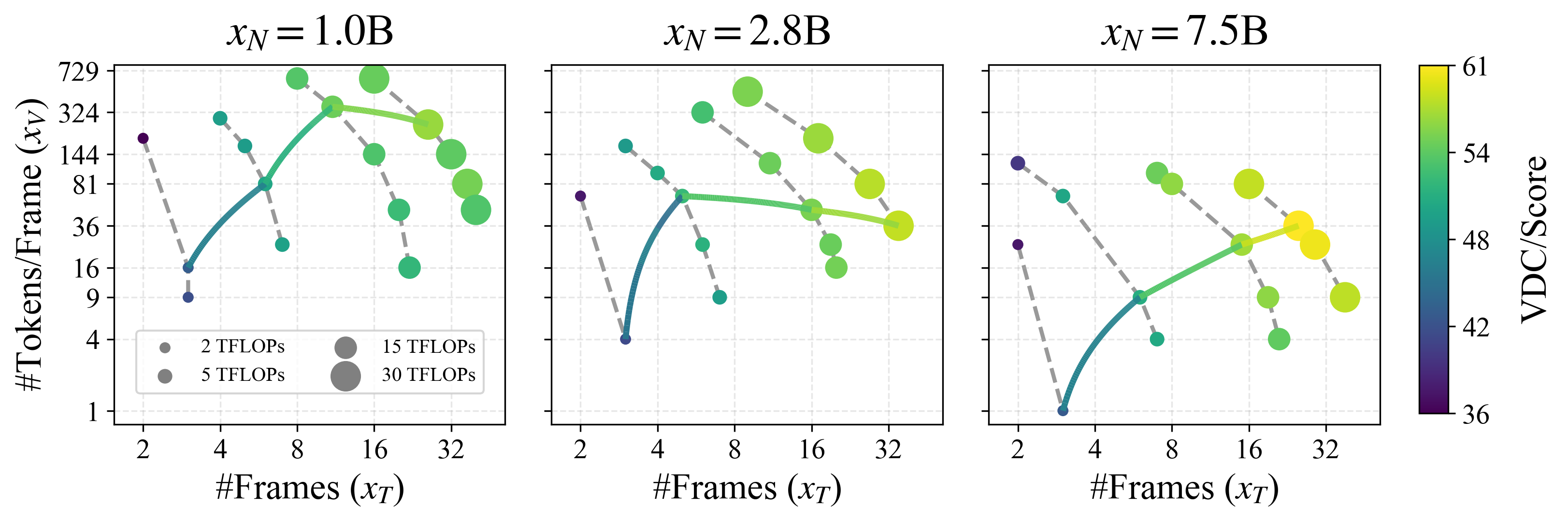}
    \myincludegraphics[width=.48\textwidth]{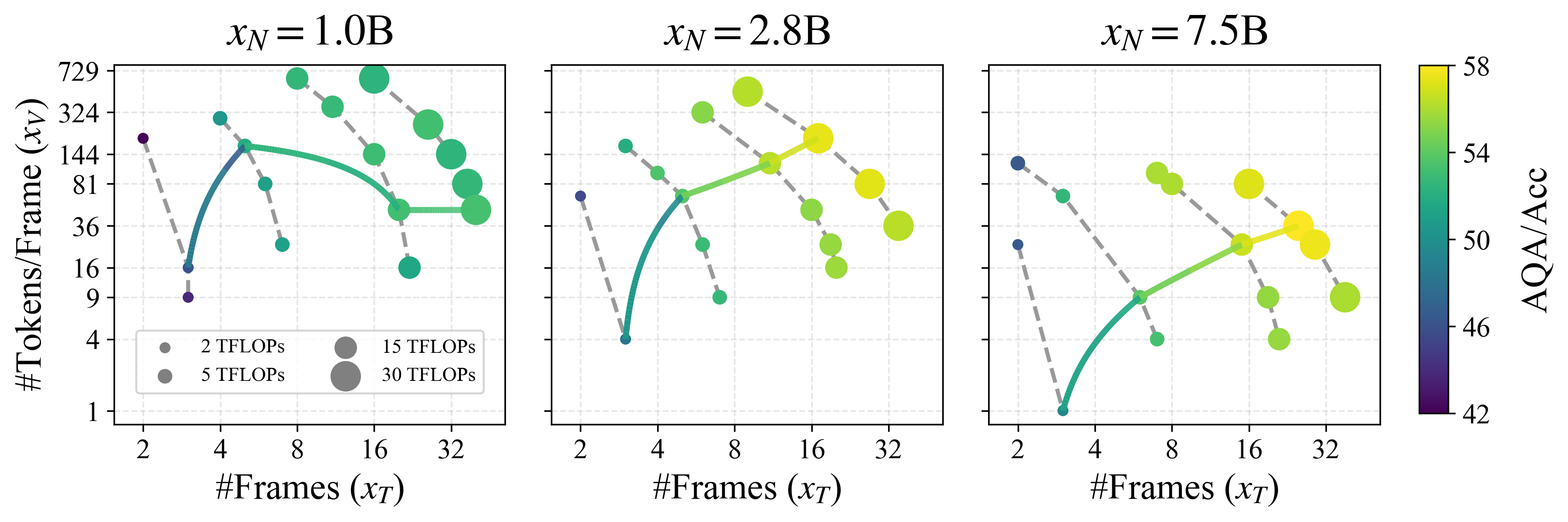}
    \myincludegraphics[width=.48\textwidth]{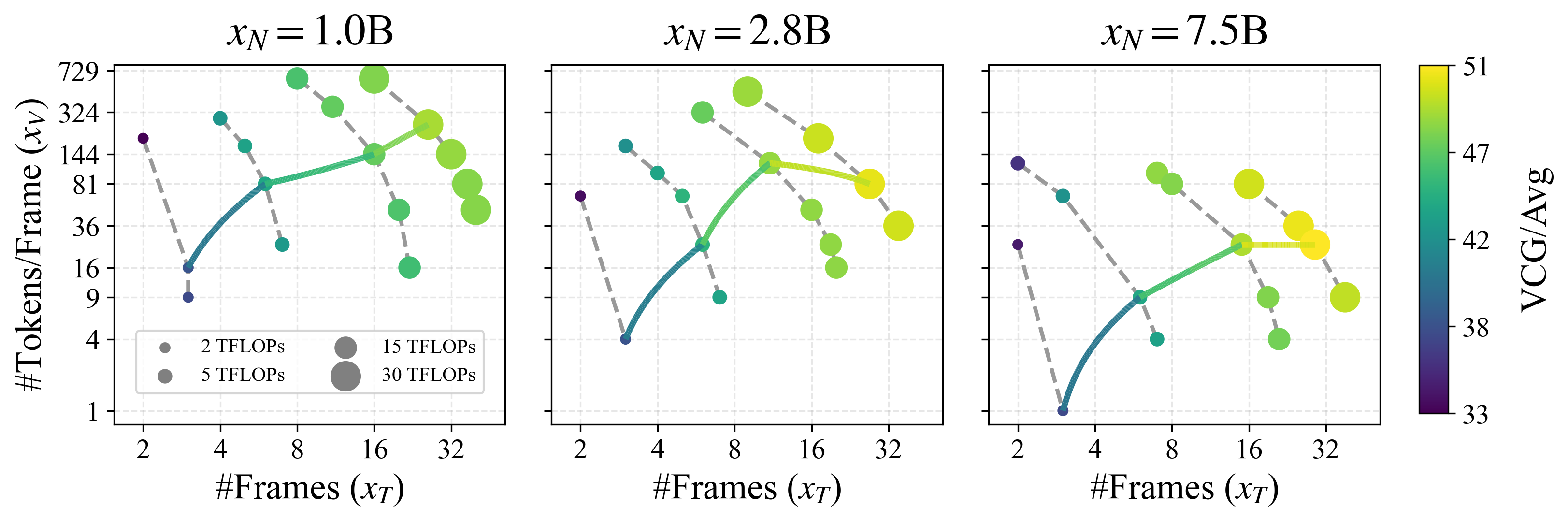}
    \myincludegraphics[width=.48\textwidth]{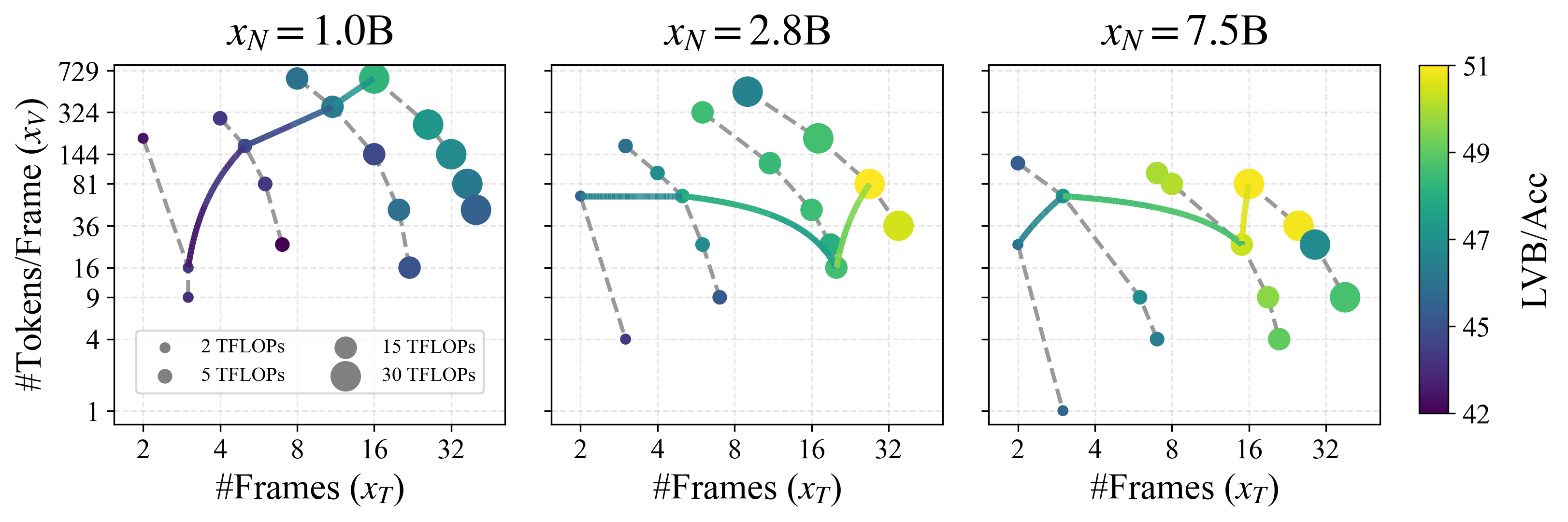}
    \myincludegraphics[width=.48\textwidth]{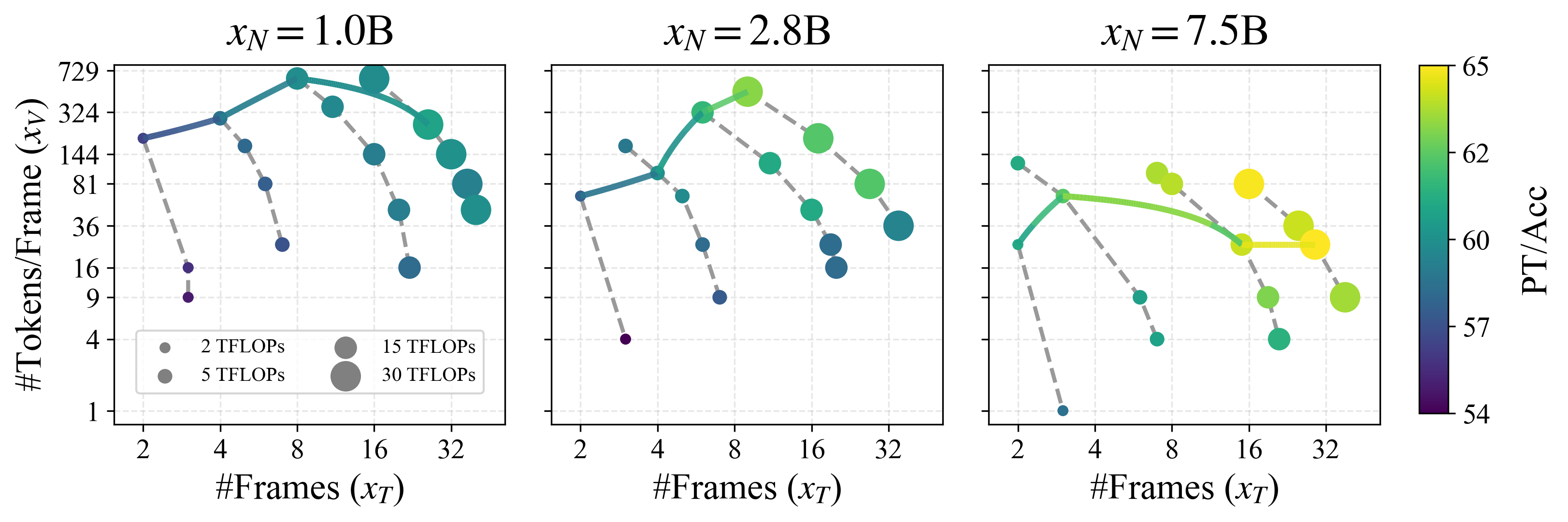}
    \myincludegraphics[width=.48\textwidth]{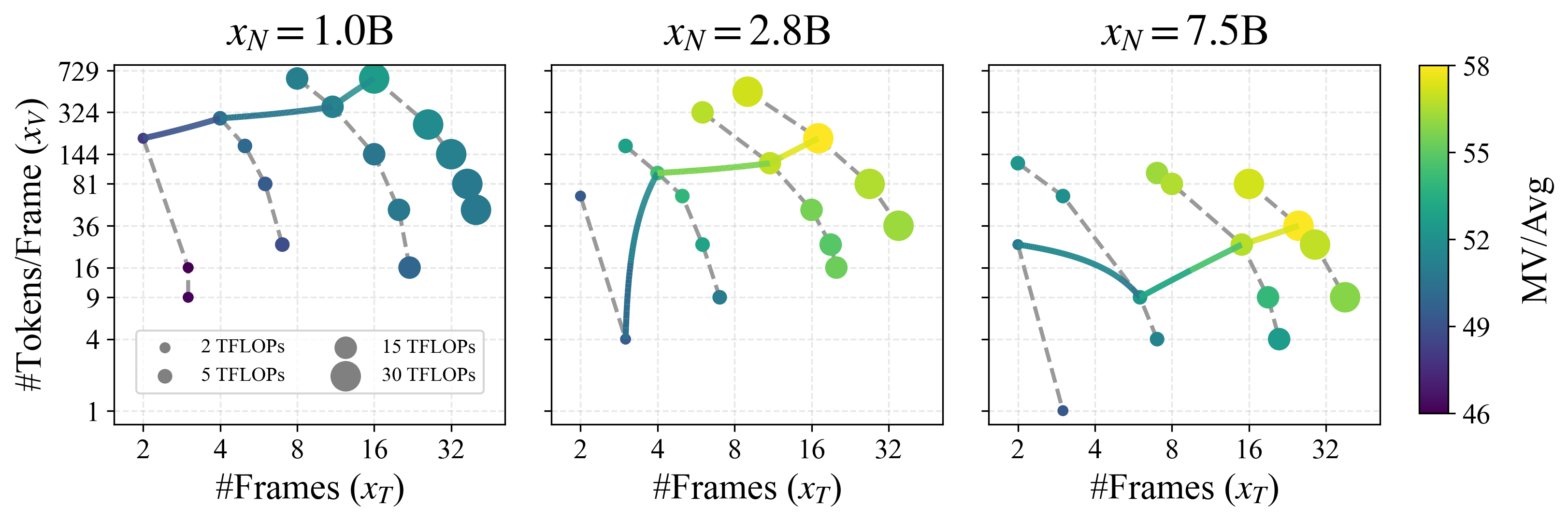}
    \myincludegraphics[width=.48\textwidth]{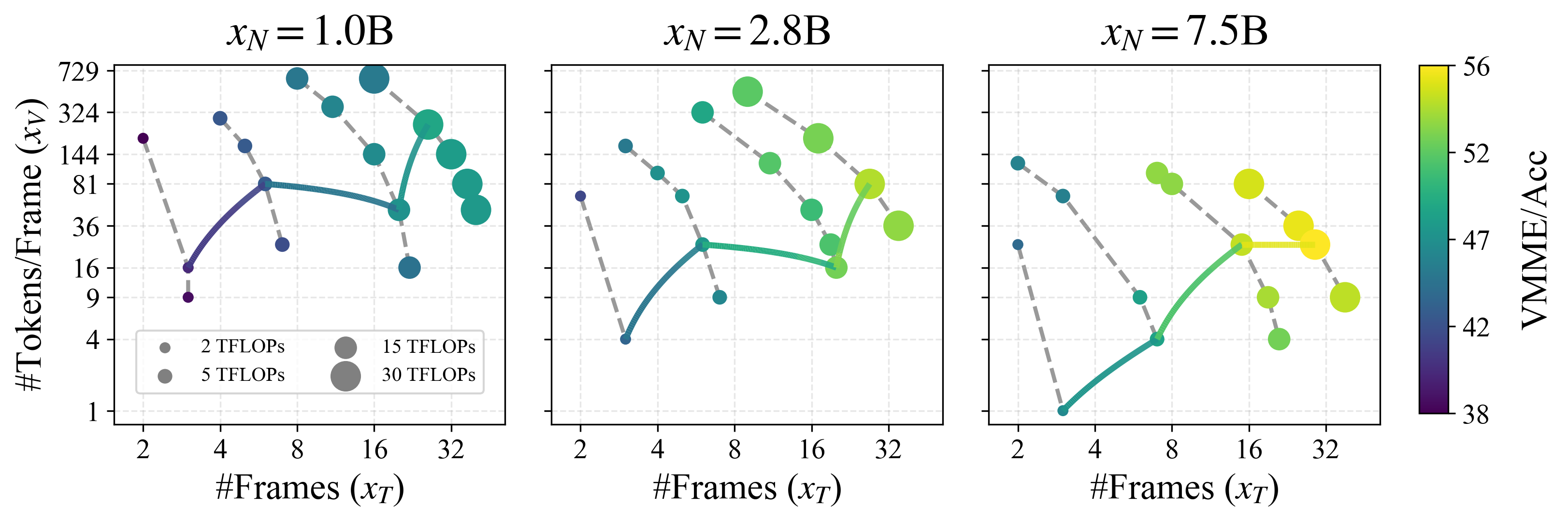}
    \myincludegraphics[width=.48\textwidth]{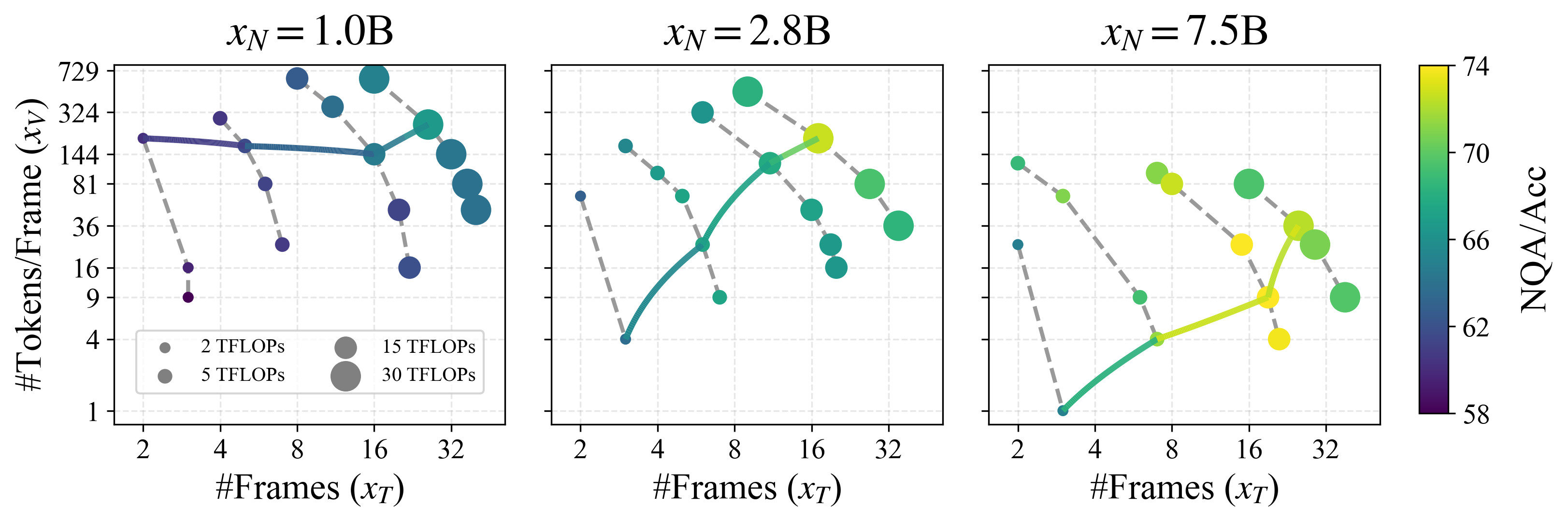}
    \caption{
      \textbf{Task-Specific isoFLOPs Curves and Compute-Optimal Frontier}. 
      }
    \label{fig:plt_isoflop_sweep_performance_trend_metrics=all}
  \end{figure*}

\end{document}